\documentclass[10pt,journal,compsoc]{IEEEtran}
\usepackage{amsmath,amsfonts}
\usepackage{algorithmic}
\usepackage{array}
\usepackage[caption=false,font=normalsize,labelfont=sf,textfont=sf]{subfig}
\usepackage{textcomp}
\usepackage{stfloats}
\usepackage{url}
\usepackage{verbatim}
\usepackage{graphicx}
\usepackage{bbm}
\usepackage[table]{xcolor}
\usepackage{setspace}
\usepackage{multirow}
\usepackage[ruled,vlined]{algorithm2e}
\usepackage{etoolbox}
\usepackage{tabularray}

\begin{document}
%
\title{Do as I Do: Pose Guided Human Motion Copy}
%
%

%

\author{Sifan Wu,
       Zhenguang Liu,
       Beibei Zhang,
       Roger Zimmermann,~\IEEEmembership{Senior~Member,~IEEE},
       Zhongjie Ba, 
       Xiaosong Zhang, 
        and~Kui Ren,~\IEEEmembership{Fellow,~IEEE}

\IEEEcompsocitemizethanks{\IEEEcompsocthanksitem 
	Sifan Wu is with the School of Computer Science and Technology, Jilin University, Changchun 130015, China (email: wusifan2021@gmail.com).\protect
	\IEEEcompsocthanksitem Zhenguang Liu (Corresponding author), Zhongjie Ba, and Kui Ren are professors of School of Cyber Science and Technology, Zhejiang University, Hangzhou 310058, China (email: liuzhenguang2008@gmail.com, \{zhongjieba,kuiren\}@zju.edu.cn).\protect%
		\IEEEcompsocthanksitem Beibei Zhang is with Zhejiang Lab, Hangzhou, Zhejiang Province, 311121, China (e-mail: bzeecs@gmail.com).
	\IEEEcompsocthanksitem Roger Zimmermann is a professor of School of Computing, National University of Singapore, 119613, Singapore (e-mail: rogerz@comp.nus.edu.sg).\protect
		\IEEEcompsocthanksitem Xiaosong Zhang is with the Center for Cyber Security, the College of Computer Science and Engineering, University of Electronic Science and Technology of China, Chengdu, Sichuan, 611731, China (e-mail: johnsonzxs@uestc.edu.cn).\protect

}

}

%

\markboth{IEEE Transactions on Dependable and Secure Computing}%
{Wu \MakeLowercase{\textit{et al.}}: Do As I Do: Pose Guided Human Motion Copy}
%

\IEEEtitleabstractindextext{%
\begin{abstract}
Human motion copy is an intriguing yet challenging task in artificial  intelligence and computer vision, which strives to generate a fake video of a target person performing the motion of a source person. The problem is inherently challenging due to the subtle human-body texture details to be generated and the temporal consistency to be considered.  Existing approaches typically adopt a conventional GAN with an L1 or L2 loss to produce the target fake video, which intrinsically necessitates a large number of training samples that are challenging to acquire.  Meanwhile, current methods still have difficulties in attaining realistic image details and temporal consistency, which unfortunately can be easily perceived by human observers.

\quad Motivated by this, we try to tackle the issues from three aspects: (1) We constrain pose-to-appearance generation with a perceptual loss and a theoretically motivated Gromov-Wasserstein loss to 
bridge the gap between pose and appearance. 
(2) We present an episodic memory module in the pose-to-appearance generation 
to propel continuous learning that helps the model learn from its past poor generations.  We also utilize geometrical cues of the face to 
optimize facial details and refine each key body part with a dedicated local GAN. 
(3)  We advocate generating the foreground in a sequence-to-sequence manner rather than a single-frame manner, explicitly enforcing temporal inconsistency.  Empirical results on five datasets, \emph{iPER, ComplexMotion, SoloDance, Fish, and Mouse datasets}, demonstrate that our method is capable of generating realistic target videos while precisely copying motion from a source video. Our method significantly outperforms state-of-the-art approaches and gains 7.2\% and 12.4\% improvements in PSNR and FID respectively. 
\end{abstract}

}

\maketitle

\IEEEdisplaynontitleabstractindextext

%
\IEEEpeerreviewmaketitle

\IEEEraisesectionheading{\section{Introduction}\label{sec:introduction}}

\IEEEPARstart{T}{he} seismic breakthrough of artificial intelligence has given rise to numerous intriguing and appealing video applications. A compelling application is to copy the motion from a source person to a target person, generating a fake video of the target person enacting the same motion as the source person. 
Motion copy empowers an untrained person to be depicted in videos dancing like a professional dancer, acting like a Kung Fu star, and playing basketball like an NBA player. Correspondingly, motion copy finds its applications in a wide spectrum of scenarios including  
animation production~\cite{visch2015viewer,mou2015creative}, augmented reality~\cite{carmigniani2011augmented,siriwardhana2021survey}, and social media entertainment~\cite{tulyakov2018mocogan}. Interestingly, the source and target persons might be greatly different in body shape, appearance, and race. 

\begin{figure}[ht]
	\centering
	\includegraphics[width=\columnwidth]{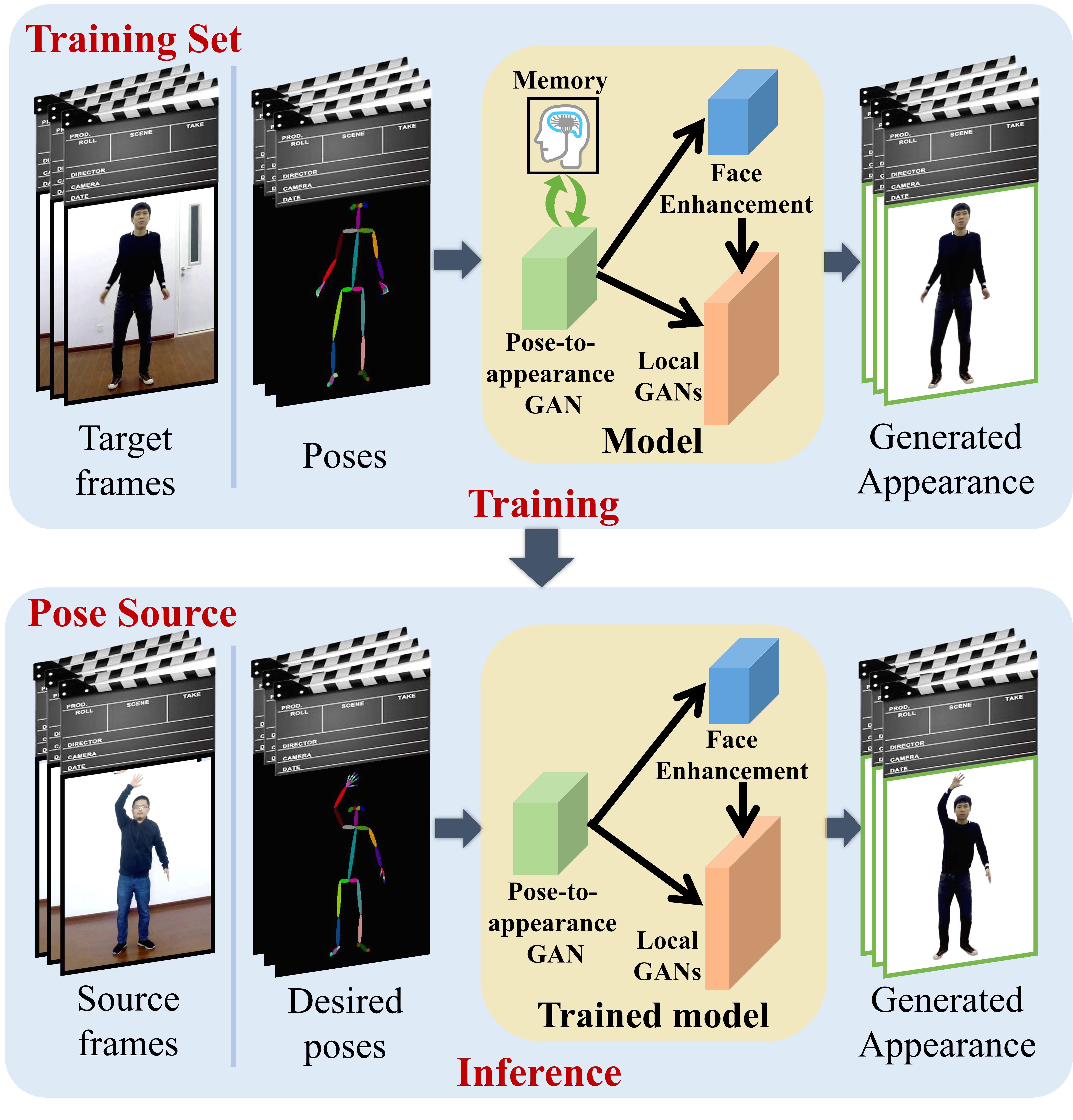}
	\caption{\textit{In the training stage}, we extract the poses from the given video frames of a target person and feed the poses into the model, which generates the video frames of the target person. \textit{In the inference stage}, we input the desired poses, which may be extracted from a video of a source person, and input them to the trained model to generate the frames of the target person.}
	\label{fig1}
\end{figure}

Fundamentally, motion copy amounts to learning a mapping from the given video of a source person to the target video of a target person, as shown in Fig.~\ref{fig1}.
The task is inherently challenging due to the high dimensionality of the mapping 
and subtle motion details to be generated. Technically, each frame of the target fake video comprises millions of pixels. Even a few wrong pixels are highly noticeable to human observers. 

Generally, motion copy is carried out in two steps. In the first step, the \emph{pose} or \emph{mesh} sequence of the source person is extracted from the source video. In the second step, motion copy learns a generative model that maps the intermediate representation (\emph{pose} or \emph{mesh} sequence) to the appearance of the target person, synthesizing the fake video where the target person enacts the motion of the source person.  
One line of works extracts human poses as the intermediate representation, 
which are referred to as \emph{pose-guided} methods~\cite{chan2019everybody, wang2019few, ghafoor2022walk}.
Another line of works captures human body meshes as the intermediate representation, 
which are termed as \emph{warping-guided} methods~\cite{liu2019liquid, wei2021c2f}. Recently, a few approaches advocate to transfer motion directly in the image feature space~\cite{joo2018generating}  or introduce neural rendering techniques to reconstruct human templates
from static images~\cite{huang2021few}.
In this paper, we focalize pose-guided target video generation, in view of its efficiency and robustness to cloth deformation.

Upon investigating and experimenting on the released implementations of 
state-of-the-art methods~\cite{chan2019everybody, balakrishnan2018synthesizing, wang2019few, liu2019liquid, wei2021c2f}, we empirically observe the following issues: 
(1) Current pose-to-appearance generation models primarily hinge on either L1 or L2 loss to train a GAN 
that bridges the gap between a pose and its target appearance.
Such GANs necessitate a large number of training samples. 
Nevertheless, we often have only one or a few videos of the target person for training. 
(2) Whereas existing methods achieve plausible results on a broad stroke, the
issues of distorted faces, hands, and feet are quite rampant. The high-fidelity textures of the face, hands, and feet, \emph{which either require sufficient details or have flexible movements}, are usually missing. 
(3) Most existing methods generate each frame independently,  
ignoring the fact that adjacent frames are closely related to each other. 
This usually leads to temporal inconsistency in the generated video. 

In this paper, we embrace three key designs to tackle the challenges. 
(1) We augment our pose-to-appearance GAN with a theoretically motivated Gromov-Wasserstein loss and a perceptual loss, 
which alleviates the problem of scarce training samples and attains realistic results. 
(2) We propose an episodic memory module in the pose-to-appearance generation so that the model  continuously accumulate experience from its past poor generations.  We also utilize geometrical cues of the face to 
optimize facial details and refine each key body part with a dedicated local GAN. 
(3)  
 We instill spatial coherency and temporal consistency into our generated video by designing a spatial-temporal discriminator.

Interestingly, most existing methods typically focus on fake \emph{human} motion generation. In this paper, we explore applying our approach to a range of objects including humans, fish, and mice. Extensive experiments are conducted on benchmark datasets including iPER, ComplexMotion, SoloDance, Fish, and Mouse datasets. Empirically, our approach outperforms state-of-the-art approaches by a large margin (7.2\% and 12.4\% gain in PSNR and FID) in fake video generation. 

To summarize, the \textbf{key contributions} of this work are:

\begin{itemize}
	\item We investigate the novel framework of incorporating Gromov-Wasserstein loss and perceptual loss for pose-to-appearance generation, which encodes  
	pairwise distance constraints and attains realistic results.
	\item  In light of the divide-and-conquer strategy \cite{liu2021aggregated}, we polish the local regions of key body parts including face, hands, and feet separately with dedicated local GANs.  We empirically present a new vector field incorporating ears to characterize the face orientation, which serves to identify  frames  with similar face orientations to enhance the generated face.
	\item Extensive experiments show that our approach achieves state-of-the-art performance. Besides, our approach could be generalized to other articulated objects, including fish and mouse.
\end{itemize}

We would like to share that this paper is the continuation of our earlier work “Copy Motion From One to Another: Fake Motion Video Generation” published in IJCAI 2022 \cite{liu2022copy}, which is accepted as a Long Presentation paper at an acceptance rate of 3.75\% (the paper acceptance rate is 15\%, Long Presentation papers are papers that rank top 25\% among the accepted papers). This work is distinct from the conference version paper in four aspects. (1) Unlike our earlier work, which generates each frame independently in a sequence-to-frame framework, this work generates $k$ consecutive foreground frames simultaneously with a sequence-to-sequence framework, encoding the wealth of temporal context information. (2) In this paper, we propose a novel episodic memory component that stores the poor generations of the model and replays these samples occasionally to enforce the model continuously learning from its defects. (3) To capture the orientation of the human face, in contrast to the mouth vector on the face employed in our earlier work, we experimentally discover that the geometric information from the ears vectors on the face is more stable and significant. Inspired by this, we present a new vector field to characterize the face orientation. (4) This work consistently outperforms the earlier work on iPER and ComplexMotion datasets and provides more insights and findings in human motion copy. Significantly, our earlier work focuses on only human motion. In this paper, we explore applying our approach to a range of objects including humans, fish, and mice.

 The remainder of the paper is organized as follows. In Section~\ref{related work}, we give a brief introduction to the related work of image synthesis and human motion copy. Thereafter, we elaborate on the proposed method in Section~\ref{method}. In Section~\ref{experiment}, we present the experiments and performance analysis. Finally, we conclude the paper in Section~\ref{conclusion}.


\section{Related Work}
\label{related work}
Before diving into the details of our approach, let us first review and categorize the related works on motion copy.  We first recap the holistic view of image synthesis, which provides a broader range of research pertinent to human motion copy. 
We then present the hitherto human motion copy approaches, which can be cast into three categories, namely \emph{pose-guided} human motion copy, \emph{warping-guided} human motion copy, and \emph{no-intermediary} human motion copy.
\subsection{Image Synthesis} Earlier research resorts to Variational Autoencoder~\cite{kingma2013auto} and Auto-Regressive models~\cite{van2016pixel} for image synthesis. Recently, the proposal and application \cite{zheng2021grip, zhang2021pise} of Generative Adversarial Networks (GANs)~\cite{goodfellow2014generative} have led to great advancement in image generation. Technically, GANs utilize a generator-discriminator architecture, where the generator produces images and the discriminator distinguishes between real and fake images. The generator and discriminator are iteratively optimized in a two-player min-max game. Conditional GANs synthesize the images under a given conditional input (\emph{e.g.}, class labels). Isola et al.~\cite{isola2017image} consider the conditional GAN as a general solution to accomplish image synthesis tasks such as image reconstruction, style transfer, and image coloring. 
Rather than generating a vanilla image,~\cite{zhang2017stackgan, zhang2021pise} propose a two-stage GAN to produce a high-resolution image. Upon initiating photo-realistic images, Gao L et al.~\cite{gao2021lightweight} propose a lightweight network structure that contains a generator and two discriminators to generate two images with different sizes in a feed-forward process. GANs have made remarkable progress in recent years on many tasks \cite{cao20193d, he2018multi, yang2019embedding, shehnepoor2021scoregan}. 
However, it is well known that GANs are difficult to train and the training process is usually unstable. Towards an easy-to-train and stable GAN, Arjovsky et al.~\cite{arjovsky2017towards, gulrajani2017improved} propose a WGAN that introduces a novel Wasserstein loss. Their proposed Wasserstein metric has a superior smoothing property compared to KL divergence of GANs, which can theoretically solve the gradient vanishing problem. A drawback of these methods lies in requiring a considerable amount of samples to train a model, which might limit their applications in human motion copy where we may not have a large number of training samples available. 

\subsection{Human Motion Copy} Existing approaches for human motion copy can be roughly categorized into three groups, \emph{namely} pose-guided, warping-guided, and no-intermediary methods.

\textbf{Pose-Guided Human Motion Copy.}\quad \cite{ma2017pose} is the first seminal work of human motion copy, which proposes a two-stage detailed generation from coarse to fine. Since then, a great deal of research has been conducted on human motion copy. 
Pumarola et al. \cite{pumarola2018unsupervised, wang2018video, bansal2018recycle} employ generators and discriminators to reconstruct the target person image with arbitrary poses. Esser et al. \cite{esser2018variational, balakrishnan2018synthesizing} propose a unique conditional U-Net, which regulates the output of the variant auto-encoder on appearance. However, these researches are extremely reliant on large-scale training samples, which is difficult to fulfill in practical applications. Ren et al. \cite{ren2020human, xu2020pose} achieve great image quality with posture augmentation and novel image refinement. %
Ghafoor et al. \cite{ghafoor2022walk} proposed a novel video-to-video action transfer framework, which consists of a cascaded sequence of action transfer block with multi-resolution structure similarity loss. Yang et al. \cite{yang2020transmomo} perform human video motion transfer in an unsupervised manner, which utilizes the invariance of three orthogonal variation factors, including motion, structure, and view. Nonetheless, these methods fail to take into account the importance of maintaining facial details during the transfer process of human motion. Although Chan et al. \cite{chan2019everybody} introduce a face enhancement module, due to the overfitting problem of GAN, it is not effective in generating satisfactory faces.  In contrast, our body parts enhancement polishes the generated face with a self-supervised training scheme and refines the key body portion using  dedicated local GANs. 

\textbf{Warping-Guided Human Motion Copy.}\quad
Dong et al.~\cite{dong2018soft, liu2019liquid, liu2021liquid} disentangle the human image into action and appearance, and then perform motion imitation by a warping GAN that distorts the image according to reference poses. Similar to the above method, Shysheya et al. \cite{shysheya2019textured, liu2020neural} introduce an attention mechanism between pose skeleton and image to generate UV coordinates and then warps patch-level human texture maps to adapt the UV coordinates. However, these methods are limited by the diversity of texture maps, resulting in blurs and artifacts of the generated video. Han et al. \cite{han2019clothflow} focus on learning an appearance flow that warps the clothing of a target person to the corresponding area of the source person. Wei et al. \cite{wei2021c2f} warp the motion of the target human image and then refine the details.  Nevertheless, the warping-based motion copy method, by nature, has difficulties in coping with rapid human motion. Moreover, these methods disregard the temporal consistency across frames, resulting in discontinuous video and visual artifacts.

\textbf{No-Intermediary Human Motion Copy.}\quad
There are also attempts that direct their efforts at  motion copy without any intermediaries (i.e., poses or meshes). Joo et al. \cite{joo2018generating} employ two specific losses to constrain the GAN which generates a fusion image (one's identity with another's shape). However, it deeply concentrates on upper body motion copy (without legs and feet) and eye style transfer. In contrast, our model does not only achieve human whole-body motion copy but also boldly tries motion copy between animals. To the best of our knowledge, we are the first to replicate movements in other articulated objects of the same species, including fish and mice.

\begin{figure*}[ht]
	\centering
	\includegraphics[width=\textwidth]{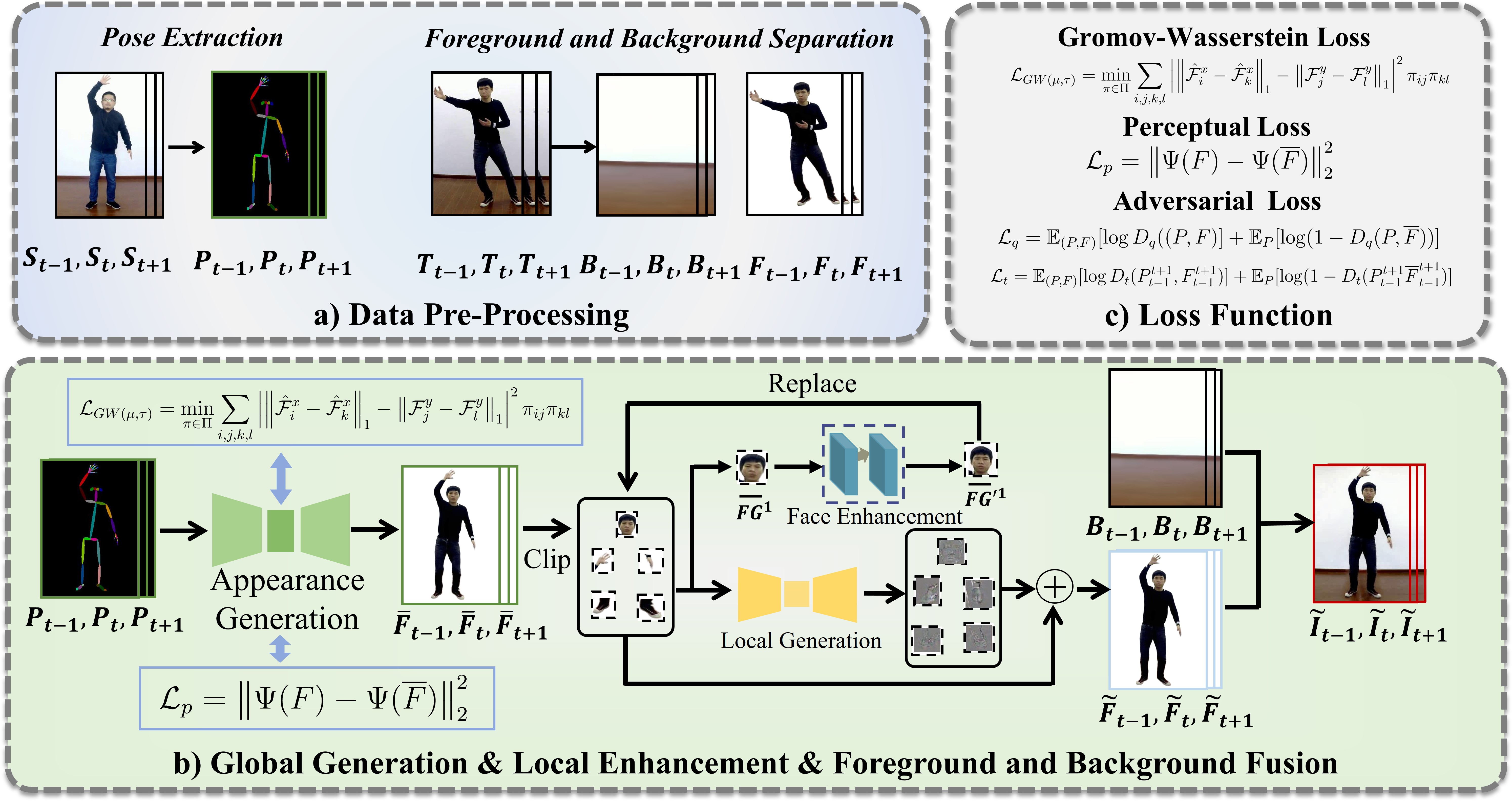}
	\caption{A high-level overview of our method. (a) For data pre-processing, we extract the pose sequence of the source video and separate the background from the target video. (b) We first generate the appearance sequence guided by the poses. Then the local body parts are enhanced with face enhancement and multi-local GANs. Finally, the refined appearance (foreground) and the background are coupled into a frame. (c) We train the appearance generator utilizing a Gromov-Wasserstein loss, a perceptual loss and an adversarial loss.	}
	
	\label{fig2}
\end{figure*}

\section{Our Method}
\label{method}
\textbf{Problem Formulation.}\quad
Broadly, given two videos, one video for the target person whose appearance we would like to synthesize and the other video for the source person whose actions we would like to copy~\cite{chan2019everybody}, we are interested in generating a fake video of the target person performing the same actions as the source person. 

\textbf{Method Overview.}\quad
An overview of our method \textit{FakeVideo} is outlined in Fig.~\ref{fig2}. Overall, \textit{FakeVideo} consists of four key components: (1) The pose extraction module draws out the human poses from the video of the {source person}, where the poses serve as motion copy intermediaries. The foreground and background separation module segments the video of the {target person} into foreground (\textit{i.e.} human body) sequence and background sequence. (2) The pose-to-appearance GAN generates an appearance sequence for the target person from the extracted pose sequence. The local enhancement module is further engaged to polish the local regions of key body parts (face, hands, and feet). (3) The episodic memory component stores the poor generations of the model and replays these samples occasionally to enforce the model continuously learns from its own defects. (4) The foreground and background fusion module generates a fake video by fusing the polished foreground sequence and the background sequence. We would like to highlight that our generator has an edge in adopting Gromov-Wasserstein and perceptual losses while being equipped with memory components. Meanwhile, our discriminator games in spatial and temporal dual constraints, driving the generator to approach better generations. In what follows, we elaborate on the four key components in detail.

\subsection{Pose Extraction and Foreground-Background Separation}
\textbf{Pose Extraction.}\quad The goal of motion copy is to learn a mapping between  a given video of the source person and  the target video of the target person. Unfortunately, each frame of the two videos has millions of pixels, making it extremely difficult to acquire the mapping directly. Inspired by the rapid development of pose estimation techniques~\cite{liu2018human, ghafoor2022quantification, aftab2022boosting}, we utilize pose skeleton sequence as the intermediary for motion copy. The pose sequence unambiguously indicates the motions and can be used to guide body appearance generation. To this end, we shift to learn a mapping from the poses to the body appearance sequence. Particularly, we adopt pre-trained pose detectors OpenPose~\cite{cao2017realtime} and DCPose \cite{liu2021deep} to extract poses from videos. 

\begin{figure*}[ht]
	\centering
	\includegraphics[width=0.85\textwidth]{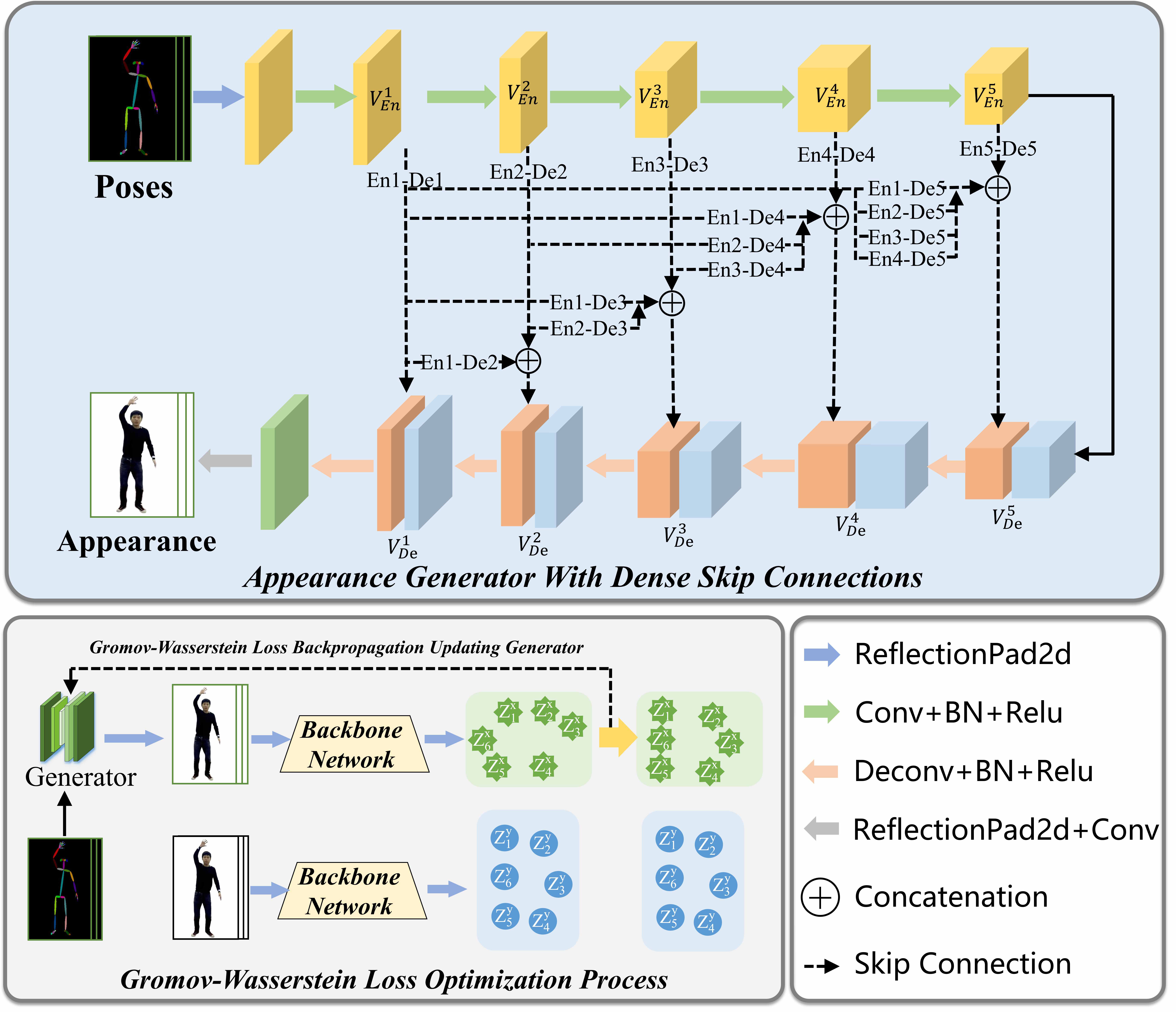}
	\caption{Pose-to-appearance generator and Gromov-Wasserstein loss. Top: We adopt an encoder-decoder architecture with dense skip connections that facilitate the fusion of features across scales. Bottom: The Gromov-Wasserstein loss is introduced to guide the pose-to-appearance generation. 
	}
	\label{fig3}
\end{figure*}

\textbf{Foreground and Background Separation.}\quad The pose skeleton clearly characterizes the motion, however, we believe it is too ambitious to synthesize a full frame (foreground and background) directly conditioned on a desired pose. Instead, an important step of our pipeline is to compute a mask matrix $M$, which is leveraged to explicitly disentangle each video frame into foreground and background. We devise a generator to concentrate on synthesizing \emph{only} the foreground sequence from poses. This facilitates our model to avoid considering a large number of background pixels in the pose-to-appearance generation, resulting in a more realistic appearance of the human and faster convergence for the network. Specifically, we adopt the off-the-shelf Mask-RCNN \cite{he2017mask} to obtain the mask matrix $M$. In addition, we employ image inpainting technology~\cite{yu2019free} to fill the removed foreground pixels in the background. 

\subsection{Pose-to-appearance Generation and Local Enhancement}
Now, we consider how to generate an appealing body foreground sequence upon a given pose sequence. Technically, we design a pose-to-appearance generation GAN (appearance GAN), consisting of a \emph{generator} that incorporates {perceptual loss and Gromov-Wasserstein loss}, and a \emph{discriminator} that exerts spatio-temporal dual constraints. 

\textbf{Dense Skip Connections in Generator.}\quad The structure of the generator is illustrated in Fig.~\ref{fig3}, where we engage in a U-shaped architecture with multiple encoder-decoder layers. In conventional U-Net, a decoder layer solely connects to one symmetry encoder layer \cite{cheng2020deepmnemonic, wang2022softpool++}. These relatively isolated relationships between different level encoder-decoder layers lead to insufficient spatial information modeling in the encoding and decoding process. Explicitly, during the encoding process of conventional U-Net architecture, consecutive convolutions in the encoder would inevitably drop some low-level detailed features. To tackle the challenge, we devise dense skip connections in the U-shaped architecture. Our motivation is to preserve rich features from multiple levels rather than using only one level feature in the foreground generation. Therefore, as shown in Fig.~\ref{fig3}, instead of connecting a decoder at layer $i$ with only the symmetric encoder at layer $i$, we add extra skip connections from the encoders at layers $\{1, 2, \cdots, i-1\}$ to the decoder at layer $i$. For example, decoder layer \emph{De-layer4} not only receives the feature information from the hop connection of encoder layer \emph{En-layer4} (as in the conventional U-Net), but also receives the feature information from encoder layers \emph{\{En-layer1, En-layer2, En-layer3\}}. In this way, each decoder could integrate multi-level latent features and is able to access lower-level features.

\textbf{Gromov-Wasserstein Loss and Perceptual Loss to Facilitate Appearance Generation.}\quad In the training phase, we extract the pose sequence and foreground sequence from the video of the target person, and train our generator network to learn how to capture the mapping function from the pose sequence to the corresponding foreground sequence of the target person.  Existing methods typically address this pose-to-appearance problem with a conventional GAN, and measure the discrepancy between the generated foreground frame and the ground truth frame via a pixel-wise L2 loss. Such approaches, by nature, require a large number of training samples to reach convergence. To alleviate this issue, we propose a Gromov-Wasserstein loss that preserves the distance-structure of the feature space instead of the conventional pixel-wise L2 loss. Particularly, the Gromov-Wasserstein loss enforces that the generated fake frames should have the same feature distance structure as their corresponding ground truth frames. Put differently, if two ground truth frames $F_i$ and $F_j$ are close to each other in the image feature space, the generated fake frames for them should also be close to each other. Conversely, if $F_i$ and $F_j$ are far apart in the image feature space, the generated fake frames for them should also be far apart. In this way, we are able to train the network in a pairwise manner, where the training samples are multiplied.  Besides the Gromov-Wasserstein loss, we also add a perceptual loss that further forces the generated frame to be consistent with the ground truth frame.

Formally, given a pose sequence $\langle P_1, P_2, \cdots, P_m \rangle$, the pose-to-appearance generation network synthesizes a foreground sequence  $\langle \overline{F}_1, \overline{F}_2, \cdots, \overline{F}_m \rangle$. Specifically, we denote the feature tensors of $\langle \overline{F}_1, \overline{F}_2, \cdots, \overline{F}_m \rangle$ as $\langle \mathcal{\hat{F}}_1,  \mathcal{\hat{F}}_2, \cdots,  \mathcal{\hat{F}}_m \rangle$, and the  feature tensors of the corresponding ground truth sequence $\langle F_1, F_2, \cdots, F_m \rangle$ as $\langle \mathcal{F}_1,  \mathcal{F}_2, \cdots,  \mathcal{F}_m \rangle$. Mathematically, 
\begin{align}
	\{\mathcal{\hat{F}}_k\}_{k=1}^m = \Phi(\{{\overline{F}}_k\}_{k=1}^m), \quad \{\mathcal{F}_k\}_{k=1}^m = \Phi(\{{F}_k\}_{k=1}^m)
\end{align}
where $\Phi(\cdot )$ represents a pre-trained feature extraction backbone network. Heuristically, we show in Fig.~\ref{fig3} that optimizing the Gromov-Wasserstein loss amounts to aligning the two groups of feature tensors so that the generated fake images preserve the distance structure of their corresponding ground truth images. We could view $\{\mathcal{\hat{F}}_k\}_{k=1}^m$ and $\{\mathcal{F}_k\}_{k=1}^m$ as discrete empirical distributions $\mu$ and $\tau$, which is given by: 
\begin{align}
	\begin{split}
		\mu = \sum_{k=1}^{m}\frac{1}{m}\delta_{\mathcal{\hat{F}}_k}, \quad \tau = \sum_{k=1}^{m}\frac{1}{m}\delta_{\mathcal{F}_k}
	\end{split}
\end{align}
where $\delta_{(\cdot)}$ represents the Dirac delta distribution. Then, the Gromov-Wasserstein loss for our model can be formulated as: 
\begin{equation} 
	\mathcal{L}_{GW(\mu, \tau)} = \min_{\pi\in \Pi }\sum _{i,j,k,l}\left | \left \| \mathcal{\hat{F}}_i - \mathcal{\hat{F}}_k  \right \|_1 - \left \| \mathcal{F}_j - \mathcal{F}_l  \right \|_1 \right |^2 \pi _{ij}\pi _{kl}
\end{equation} 
where $\Pi$ denotes a collection of point distributions with margins $\mu$ and $\tau$. The optimal transport matrix $\pi$ could be calculated by minimizing the square distance with L1 costs in the intra-space.

Inspired by \cite{peyre2016gromov,wu2021dual}, an entropy regularization term is introduced to ensure tractability and reversible backpropagation in the optimal transport loss optimization. In addition, we utilize the Sinkhorn algorithm and the projected gradient descent method \cite{peyre2016gromov} to solve the entropy-regularized Gromov-Wasserstein loss. Technically, the process of optimizing Gromov-Wasserstein loss is outlined in Algorithm~\ref{alg:algorithm1}.

\begin{algorithm}[h]
	\small
	\setstretch{0.88}
	\caption{Optimizing the Gromov-Wasserstein loss for pose-to-appearance generation network}
	\begin{algorithmic}[1]
		\label{alg:algorithm1}
		\STATE \textbf{Input:}\quad (i) generated feature tensors $\{\mathcal{\hat{F}}_k\}_{k=1}^m = \Phi(\{{\overline{F}}_k\}_{k=1}^m)$ and (ii) ground truth feature tensors $\{\mathcal{F}_k\}_{k=1}^m = \Phi(\{{F}_k\}_{k=1}^m)$
		
		\STATE \textbf{Output:}\quad{Gromov-Wasserstein distance $GW_\lambda$}  
		\STATE \textbf{Hyperparameters:} 	$\lambda > 0 $, projection iterations \textit{P}, Sinkhorn iterations \textit{S}
		
		\STATE \textbf{Initialize:} $\pi_{kl}^{(0)} = \frac{1}{n}, \forall k,l $, $m = j - i$ 
		\STATE \quad Cost matrix for generated feature tensors $D_{ij} = \mathcal{L}_1 (\mathcal{\hat{F}}_i,\mathcal{\hat{F}}_j)$
		
		\STATE \quad Cost matrix for ground truth feature tensors $E_{ij} = \mathcal{L}_1 (\mathcal{F}_i, \mathcal{F}_j)$
		\FOR{t = 1:P}
		\STATE initialize a tree $T_{i}$ with only a leaf (the root);\
		\STATE $C = \frac{1}{m}E^{2} \mathbbm{1}_m\mathbbm{1}_{m}^{T}+\frac{1}{m}\mathbbm{1}_m\mathbbm{1}_{m}^{T}D^2-2E\pi^{(t-1)}D^T$;
		\STATE $K = e^{(-C/\lambda)}$;
		\STATE $b^{(0)} = \mathbbm{1}_m$;
		\FOR{l = 1:S}
		\STATE $a^{(l)} = \mathbbm{1}_m \oslash  Kb^{(l-1)}$;  
		\STATE $b^{(l)} = \mathbbm{1}_m \oslash  K^T a^{(l)}$;
		\STATE \#\quad$\oslash$ defines component-wise division
		\ENDFOR
		\STATE $\pi ^{(t)} = diag(a^{(S)}) K diag(b^{(S)})$;
		\ENDFOR
		\STATE $GW_\lambda = \sum_{i,j,k,l}^{}\left\| E_{ik} - D_{jl}\right\|^2\pi_{ij}^{(P)}\pi_{kl}^{(P)}$
	\end{algorithmic}
\end{algorithm}

\textbf{Perceptual Loss.} \quad While the Gromov-Wasserstein loss facilitates the appearance generation in the presence of sparse training samples, another loss is introduced into the network to better maintain image reconstruction details. An intuitive approach is to utilize the mean squared error (MSE) loss to minimize the pixel-wise loss between the generated human foreground $\overline{F}$ and the ground truth $F$:
\begin{align}
	\begin{split}
		\mathcal{L}_{MSE} = \left\| F - \overline{F} \right\|^2_2,
	\end{split}
\end{align}
where $\left\| \cdot\right\|_2$ represents L2 norm. Nevertheless, MSE loss may produce blurry and distorted images or lead to ill-posed details \cite{2017Low}. Given this context, we adopt a perceptual reconstruction loss that constrains the generated $\overline{F}$ to approach ground-truth in the feature space:
\begin{align}
	\begin{split}
		\mathcal{L}_p &= \left \| \Psi (F) - \Psi (\overline{F}) \right \|^2_2,
	\end{split}
\end{align}
where $\Psi(\cdot) $ represents a feature extraction network. Pixel-wise loss concentrates too much on the brightness of each pixel, while feature level loss considers more on the spatial consistency. Collectively, the Gromov-Wasserstein loss and perceptual loss together facilitate appearance generation. 

\textbf{Discriminator in Pose-to-Appearance Generation.} \quad  (1) Recalling previous approaches for motion copy, they typically employ a spatial discriminator that concentrates on the quality of each frame and fails to explicitly consider video continuity.  (2) When we watch videos, we tend to take care of the quality of frames and continuity across frames. We believe it is crucial to jointly take into account spatial consistency and temporal continuity. Based on the two observations and heuristics above, we present a spatial-temporal dual constraint, consisting of a quality discriminator $D_q$ and a temporal discriminator $D_t$. 
Specifically, (1) the quality discriminator $D_q$ enforces the forged foreground image to approach the ground truth. (2) the temporal discriminator $D_t$ captures the temporal information across frames using a set of parallel dilation convolutions. $D_q$ takes  $(P_i,{F}_i)$ or $(P_i,{\overline{{F}}}_i)$ as the input while $D_t$ absorbs  $(P_{t-1}^{t+1},{F}_{t-1}^{t+1})$ or $(P_{t-1}^{t+1},{\overline{{F}}}_{t-1}^{t+1})$. Note that $P_i$ stands for pose of the $i^{th}$ frame and $\overline{{F}}_i$ denotes the generated foreground for $P_i$. $P_{t-1}^{t+1}$ and ${\overline{{F}}}_{t-1}^{t+1}$ represent $\langle P_{t-1}, P_t, P_{t+1} \rangle$ and $\langle  \overline{{F}}_{t-1}, \overline{{F}}_t, \overline{{F}}_{t+1} \rangle$, respectively.  Both of the two discriminators are trained to output binary labels, \emph{real} or \emph{fake}. Overall, the generator strives to create more lifelike videos to fool the dual discriminator, while the discriminator tries its best to distinguish between generated video and the ground truth. Model performance is iteratively optimized in a two-player min-max game fashion. 

The image quality often appears imperfect when the fine-grained local details are missing. We scrutinized and implemented state-of-the-art methods following their released code and parameter settings \cite{chan2019everybody, balakrishnan2018synthesizing, wang2019few, liu2019liquid, wei2021c2f}. A significant insight we gain from the experiments is that current methods still have difficulties in generating \emph{detailed} face, \emph{natural} hands, and \emph{clear} feet.  After obtaining the initial body appearance using the proposed pose-to-appearance GAN network, we further employ a self-supervised face enhancement component and multiple local GANs to polish the details of local parts.

\begin{figure}[t]
	\centering
	\includegraphics[width=\columnwidth]{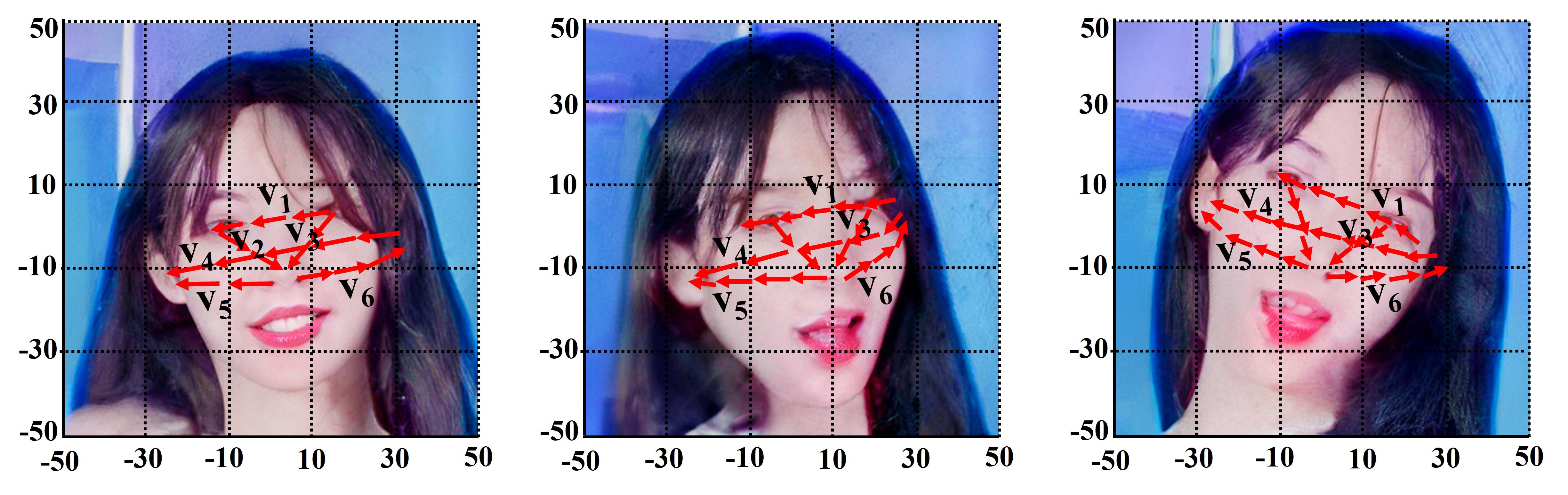}
	\caption{Face orientation is extracted from the face vector field. Three different face orientations are presented in the figure. Specifically, we employ six vectors, \emph{including $v_{1}$: right eye $\rightarrow$ left eye, $v_2$: left eye $\rightarrow$ nose, $v_3$: right eye$\rightarrow$ nose, $v_4$: right ear$\rightarrow$ left ear, $v_5$:nose$\rightarrow$ right ear, $v_6$: nose$\rightarrow$ left ear}, to characterize the face orientation.
		\label{facevector}
	}
\end{figure}

\textbf{Self Supervised Face Enhancement with Vector Field.}\quad  Intuitively, the face images of the same person with similar face orientations should look similar to each other. Therefore, we search from the given videos of the target person to identify face images that have similar face orientations as  the synthesized image. In particular, we choose multiple images with the closest face orientations rather than using only one image that has the closest face orientation as the synthesized image, making it more robust to noise.   

\begin{figure}[t]
	\centering
	\includegraphics[width=\columnwidth]{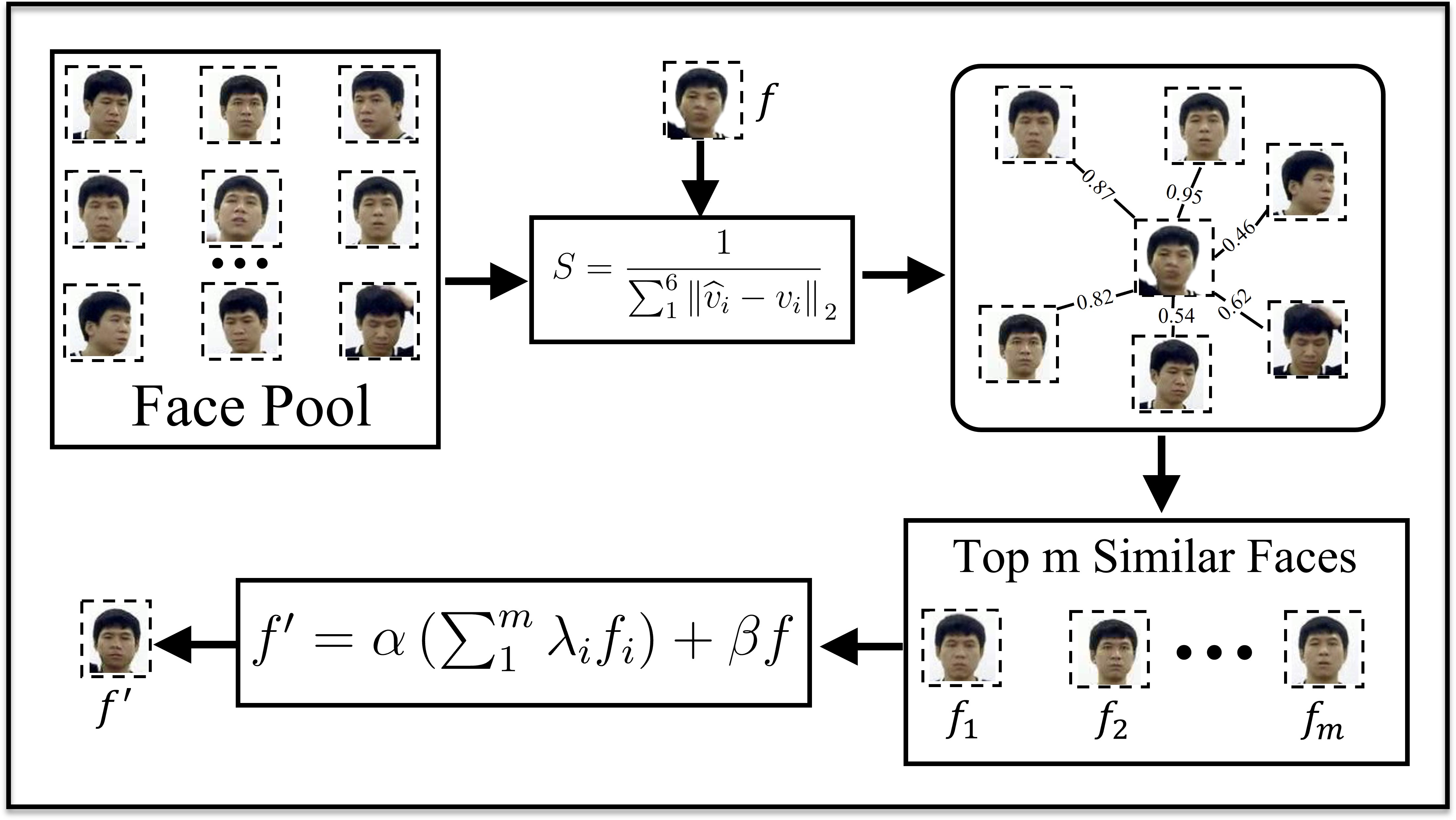}
	\caption{Self supervised face enhancement with vector field. We select $m$ faces from the target face pool with the most similar face orientation to $f$. The selected similar faces are expected to facilitate detailed face texture generation.
	}
	\label{facerefiner}
\end{figure}

For the measurement of face orientation similarity, an intuitive approach is to compute the similarity between facial features. However, facial features usually convey too much information irrelevant to face orientation, \emph{e.g., color and eye shape}.  To tackle the problem, a viable method, as shown in Fig.~\ref{facevector}, is to represent the face orientation with a face vector field. As shown in Fig.~\ref{facevector},  we employ six vectors, including $v_{1}$: right eye $\rightarrow$ left eye, $v_2$: left eye $\rightarrow$ nose, $v_3$: right eye$\rightarrow$ nose, $v_4$: right ear$\rightarrow$ left ear, $v_5$: nose$\rightarrow$ right ear, $v_6$: nose$\rightarrow$ left ear. Given two face orientations $\{v_i\}_{i=1}^6$ and $\{\hat{v}_i\}_{i=1}^6$, their similarity can be conveniently computed as:

\begin{align}
	\mathcal{S}=\frac{1} {\sum_{i=1}^6\|\hat{v}_i-v_i\|_2}
\end{align}
Subsequently, we choose top $m$ real facial images $\mathbf{f}~=~\lbrace{f_1, f_2, \dots, f_m}\rbrace$ with the largest similarity $\mathcal{S}_m$ as auxiliary faces. Finally, the generated face $f$ is enhanced into:
\begin{align}
	f^\prime = \alpha (\sum_{i=1}^m( \frac{S_i}{\sum_{j=1}^m S_j} \times  f_i)+\beta f
\end{align}
where $\frac{S_i}{\sum_{j=1}^m S_j}$ measures the weight of the $i^{th}$ chosen face $f_i$, $\alpha$ and $\beta$  are hyperparameters. The process is also depicted in Fig.~\ref{facerefiner}.
\par

\textbf{Multi-Local GANs.}\quad After enhancing the face, we further refine the face and limbs using multiple local GANs.  In light of the divide-and-conquer strategy \cite{liu2021aggregated}, we design multi-local GANs to refine key parts separately.  Concretely, we clip the five key parts $\overline{F}^i$ (face, two hands, and two feet) from the generated foreground image $\overline{F}$. We feed them into corresponding delicate GANs, which outputs a residual image ${\hat{F}_{r}}^i$, which learns the difference between the generated image of the body part and the ground truth (in terms of color and texture). Those residual images are added to $\overline{F}$ (the original foreground generation) to produce the final foreground:
\begin{align}
	\widetilde{F}^i={\hat{F}_{r}}^i+ \overline{F}^i
\end{align}

\subsection{Episodic Memory for Experience Replay}
For the pose-to-appearance generation, we adopt the theoretical Gromov-Wasserstein loss to mitigate the issue of insufficient training samples. Furthermore, inspired by lifelong learning, we introduce an episodic memory component for appearance generation, which propels continuous learning and the accumulation of past knowledge over a lifetime. More specifically, we store previous poor generations in the episodic memory and replay these poor generations periodically in training. This enforces the network to be able to consistently learn from its own mistakes and accumulate experiences. Interestingly, the mechanism is similar to our human brain that occasionally recaps significant moments recorded in our memory. The entire procedure of memory replay is formulated in Algorithm~\ref{memory}. We may describe the high-level idea as:

\begin{algorithm}[ht]
	\small
	\setstretch{0.88}
	\caption{Pose-to-appearance generation network with episodic memory component}
	\label{memory}
	\begin{algorithmic}[1]
		\STATE \textbf{Training}
		
		\STATE \textbf{Input:}\quad{training samples $\left< P_t,F_t \right>_{t=1}^T$, replay time interval K}
		
		\STATE \#\quad ${P}_i$ stands for the desired pose for the target person in the $i^{th}$ frame, and ${F}_i$ represents the corresponding appearance
		
		\STATE \textbf{Output:}\quad{generation model \textbf{G}}
		\FOR{epoch = 1:N}
		\IF { epoch mod K = 0}
		\STATE {Sample m examples from \textbf{M}} 
		\STATE \# M represents memory
		\STATE {Calculate Gromov-Wasserstein loss and perceptual loss, and then perform backpropagation to update the parameters of \textbf{G}}
		\STATE \# Experience Replay
		\ENDIF
		\FOR{t = 1:T}
		\STATE Retrieve training samples $\left< P_{t-1}^{t+1},F_{t-1}^{t+1} \right>$
		\STATE {Calculate Gromov-Wasserstein loss and perceptual loss, and then perform backpropagation to update the parameters of \textbf{G}}
		\IF {store memory}
		\STATE {Write $\left< P_{t-1}^{t+1},F_{t-1}^{t+1} \right>$ to memory \textbf{M}}
		\ENDIF
		\IF {$perceptual\_loss>loss\_threshold$}
		\STATE Write the poor generation examples $\left< P_{t-1}^{t+1},F_{t-1}^{t+1} \right>$ into memory \textbf{M}
		\ENDIF
		\ENDFOR
		\ENDFOR
		\STATE \textbf{Return} G
		\STATE \textbf{Inference}
		\STATE \textbf{Input:}\quad{the poses $P_{t=1}^T$ of source person $\mathbb{S}$, the generation model \textbf{G}}
		
		\STATE \textbf{Output:}\quad{the foreground  $\overline{F}_{t=1}^T$}
		
		\STATE \#\quad{the generated foreground (appearance) $\overline{F}$ for target person $\mathbb{T}$}
		\FOR{t in range(1:T:3)}
		\STATE $\overline{F}_{t}^{t+2}$ = \textbf{G} ($P_{t}^{t+2}$);
		\ENDFOR
		\STATE \textbf{return $\overline{F}_{t=1}^T$}
	\end{algorithmic}
\end{algorithm}

In the first epoch, we utilize all training samples to train the pose-to-appearance generation network (Algorithm \ref{memory} lines 12-14). We then select all the poorly generated samples (when the perceptual loss exceeds a threshold) and randomly select a few other samples, and put them into the episodic memory (Algorithm \ref{memory} lines 15-19). In the following epochs, we retrain all the training samples for the second time. During the second-time  training, we randomly select several samples from the episodic memory and replay (retrain) them to update the parameters of the pose-to-appearance generation network per K epochs (Algorithm \ref{memory} lines 6-11). 
Memory replay would keep the model from catastrophic forgetting, continuously improving the generated frames by learning from past poor generations. However, overfitting problems may arise if the training samples in the memory are revisited too frequently. Following \cite{de2019episodic}, our memory replay is designed to be executed only occasionally.

\subsection{Foreground and Background Fusion}

Up to this point, we have obtained the polished foreground~$\widetilde{F}$. In the pre-processing phase, we have computed the mask matrix $M$ of the foreground in the image and have refilled foreground pixels in the background~$B$ following~\cite{yu2019free}. We utilize a linear sum to couple foreground~$\widetilde{F}$ and background~$B$.
\begin{align}		
	\tilde{I} = M \odot \widetilde{F} + (1-M) \odot B
\end{align}
\subsection{Loss Functions}   
For the pose-to-appearance generator, we utilize the Gromov-Wasserstein and perceptual losses.  Now, we zoom in  the loss functions of the discriminator. We introduce a standard adversarial loss, where a quality discriminator $D_q$ attempts to discern the real and generated frames:
\begin{align}
	\begin{split}
		\mathcal{L}_q&= \mathbb{E}_{({P, F})}[\log D_q((P,F)]\\ 
		&+ \mathbb{E}_P[\log(1-D_q(P, \overline{F}))]
	\end{split}		
\end{align} 

We additionally propose a temporal consistency loss to ensure the temporal smoothness of the generated video: 
\begin{align}
	\begin{split}
		&\mathcal{L}_t = \mathbb{E}_{({P, F})}[\log D_t(P_{t-1}^{t+1}, {F}_{t-1}^{t+1})]\\
		& + \mathbb{E}_P[\log(1-D_t(P_{t-1}^{t+1},{\overline{{F}}}_{t-1}^{t+1})] 
	\end{split}
\end{align}
where $D_t$ is a temporal discriminator which tries to distinguish the real frame sequence $ {F}_{t-1}^{t+1}$ and fake sequence ${\overline{{F}}}_{t-1}^{t+1}$.

\begin{figure*}[ht]
	\centering
	\includegraphics[width=0.9\textwidth]{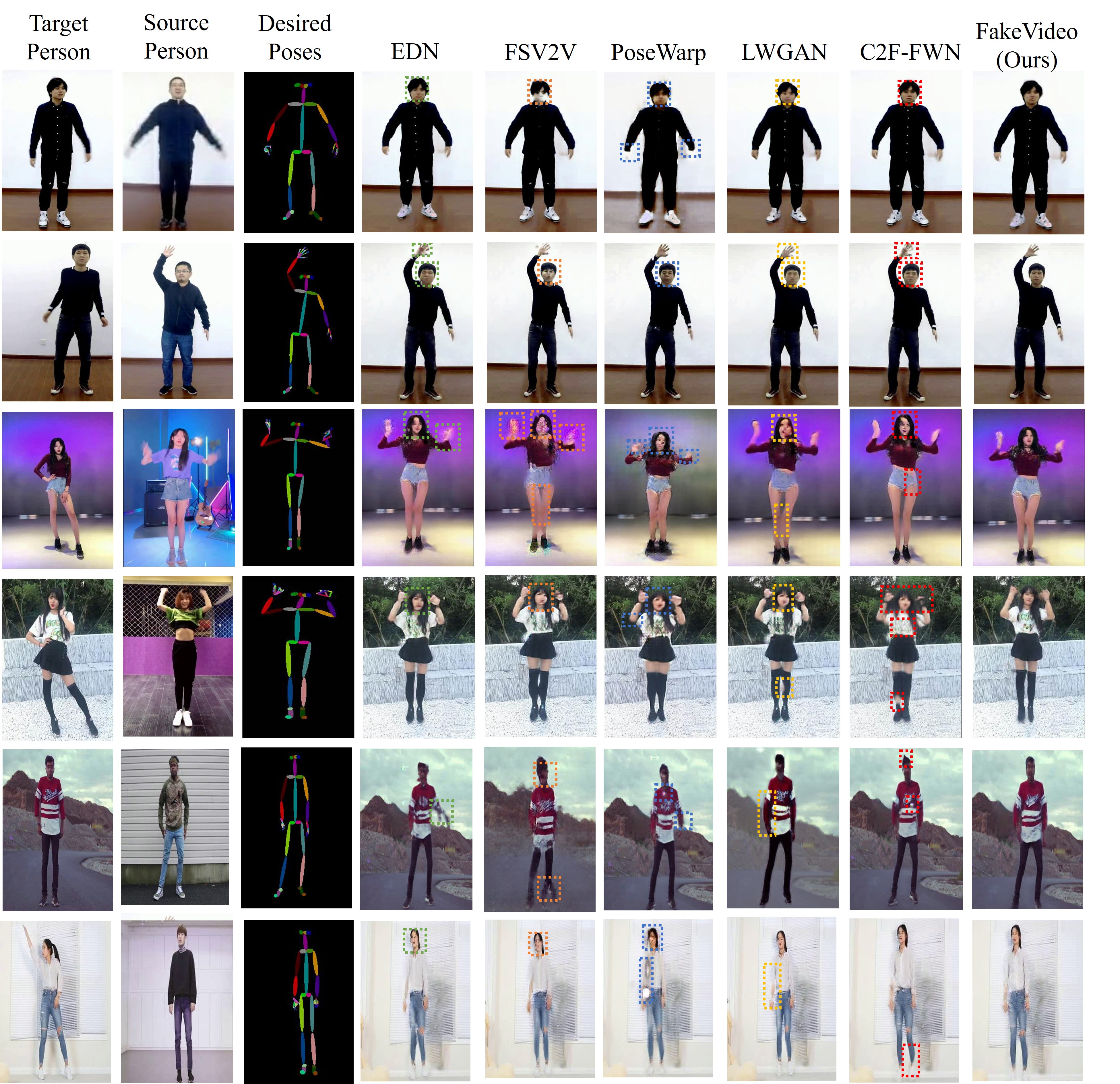}
	\caption{Visualization of the human motion copy results of different methods on three datasets. The data of the first and second rows are from the iPER dataset, the data of the third and fourth rows are from the ComplexMotion dataset, and the data of the fifth and sixth rows are from the SoloDance dataset. Columns from left to right represent the given images of the target persons, the images of the source persons, the desired poses, results of EDN, results of  FSV2V, results of  PoseWarp, results of  LWGAN, results of  C2F-FWN, and results of FakeVideo (our method), respectively. Poorly generated body parts are highlighted with dotted rectangles. Please zoom in to see more details.}
	\label{fig7}
\end{figure*}
\begin{figure*}[ht]
	\centering
	\includegraphics[width=0.9\textwidth]{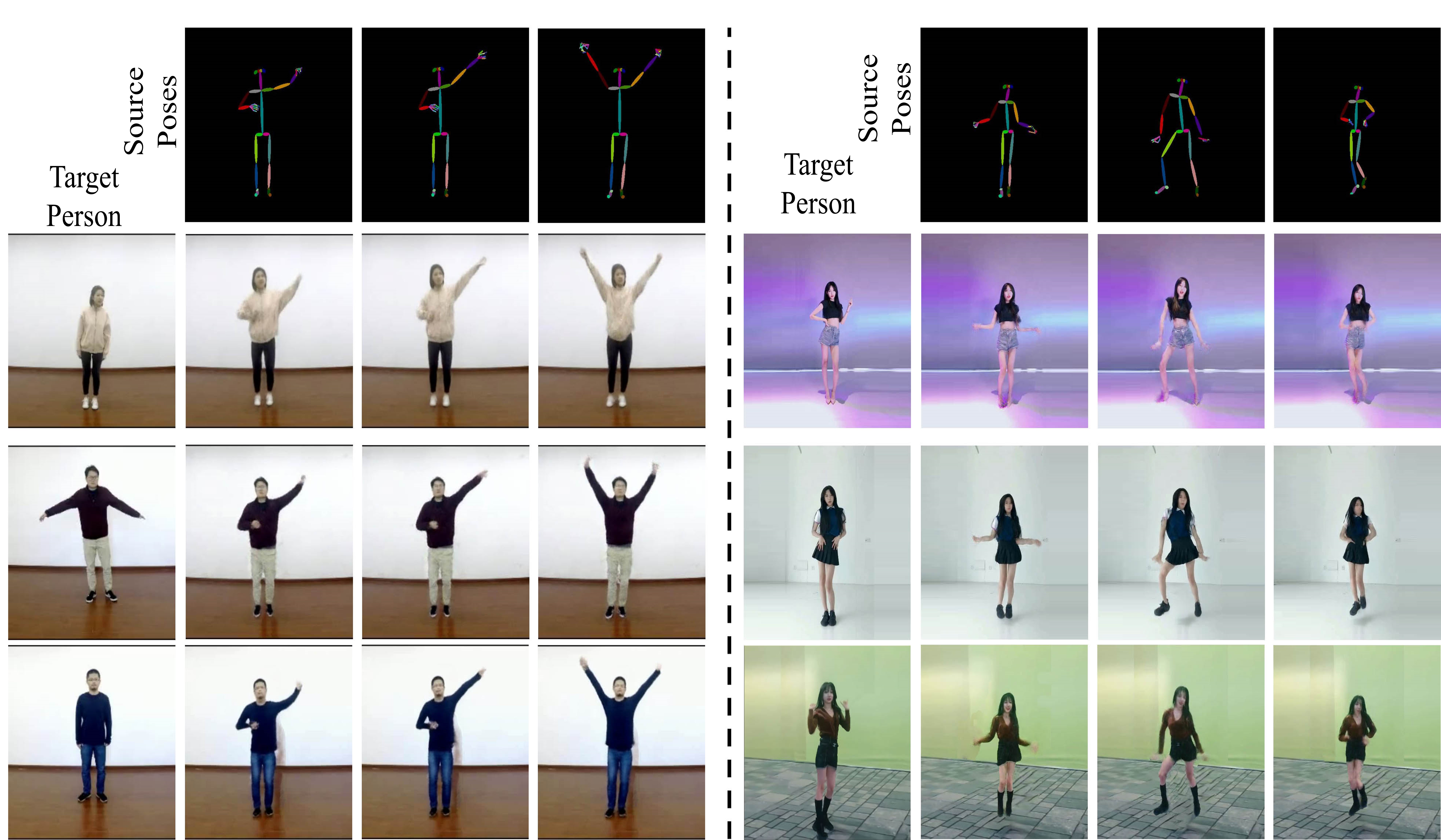}
	\caption{More visual results  on iPER and ComplexMotion datasets. }
	\label{fig6}
\end{figure*}

\section{Experiments}
\label{experiment}
In this section, we first present the experimental settings and the details of the five benchmark datasets, \emph{iPER} \cite{liu2019liquid}, \emph{ComplexMotion} \cite{liu2022copy}, SoloDance \cite{wei2021c2f} \textit{Fish dataset} \cite{xu2017lie}, and \textit{Mouse dataset} \cite{liu2022investigating}. Then, we introduce evaluation metrics for motion copy and compare our method with the state-of-the-art approaches. Further, we investigate the effects of different components in our framework. Finally, we try adapting our method to other articulated objects including fish and mice.
Briefly, we seek to answer the following research questions.
\begin{itemize}
	\item \textbf{RQ1:} How is the proposed method compared to state-of-the-art methods on human motion copy?
	\item \textbf{RQ2:} Is our method able to synthesize motion videos with attractive details for a target person? 
	\item \textbf{RQ3:} How much do different components of our method contribute to the performance? 
	\item \textbf{RQ4:} How well does the proposed method generalize to animals such as fish and mouse? 
\end{itemize}
Next, we introduce the experimental settings and empirically investigate the research questions one by one.
\subsection{Experimental Settings}
\subsubsection{Datasets} Experiments are conducted on five benchmark datasets, \emph{iPER} \cite{liu2019liquid}, \emph{ComplexMotion} \cite{liu2022copy}, SoloDance \cite{wei2021c2f}, \emph{Fish dataset} \cite{xu2017lie}, and \emph{Mouse dataset} \cite{liu2022investigating}. 

\textbf{iPER}. For human motion copy, we experiment on \textit{iPER} \cite{liu2019liquid} dataset, which contains 30 persons with different shapes, heights, and genders. A person may wear different outfits, and there are 103 outfits in total. The dataset contains 241,564 frames from 206 videos. Within the videos, different actions including 
\textit{arm exercise}, \textit{stretching exercise}, \textit{standing and reaching}, \textit{leaping}, \textit{swimming}, \textit{taichi},\textit{ chest mobility exercise}, \textit{leg stretching}, \textit{squat}, and \textit{leg-raising} are involved.

\textbf{ComplexMotion}. We also conduct experiments on \textit{ComplexMotion} \cite{liu2022copy} dataset, which contains rapid and complex motions of more than $50$ persons.
The videos are collected from various video platforms such as Tiktok\footnote{https://www.tiktok.com} and Youtube\footnote{https://www.youtube.com}.
In particular, ComplexMotion consists of 68,320 frames from 122 videos. Within the videos, persons wear various clothes and perform complex movements such as \textit{street dance}, \textit{sports}, and \textit{kung fu}.

	\textbf{SoloDance Dataset}\quad We further conduct experiments on SoloDance \cite{wei2021c2f} dataset, which contains 179 dance videos with 53,700 frames. Specifically, 143 human subjects were captured with each wearing different clothes and performing complex dances (\textit{e.g., modern, street dances}) in various scenes.

\textbf{Fish dataset}. For motion copy from a fish to another fish, we utilize the \textit{Fish dataset} \cite{xu2017lie}, which contains 14 fish videos of 6 different fishes. Specifically, each video consists of 2,250  to  24,000 frames.

\textbf{Mouse dataset}. For mouse motion copy, we use the \textit{Mouse dataset} \cite{liu2022investigating}, which includes 12 mouse videos of 4 mice. Mouse depth images were captured at 25 FPS with the top-view Primesense Carmine camera. The 3D poses of the mouse are extracted from the depth image using annotation tool of \cite{liu2022investigating}. Then, we project the 3D poses to the 2D plane, and obtain the 2D poses of the mice. Specifically, the number of frames in each video varies from 500  to 30,000.

\subsubsection{Implementation details}
We utilize PyTorch 1.4.0 to realize our proposed framework. We train the \textit{FakeVideo} framework independently on iPER and ComplexMotion. During training, all the frames are resized to 512 $\times$ 512. We utilize OpenPose to detect 18 human joints in each frame of the dataset. For \textit{Fish } and \textit{Mouse} datasets, we would like to point out that we do not further enhance the local details of the generated fish and mice, since they are small in body size and the pose-to-appearance generation seems to be sufficient to yield realistic fish and mice. We employ Mask-RCNN to disentangle the foreground sequence (body) and the background sequence from a video. We utilize the pre-trained VGGNet \cite{simonyan2014very} as our frame  feature extractor, which consists of 16 convolutional layers and 3 fully connected layers. The output of the $16^{th}$ convolutional layer is the extracted feature, which is used for perceptual loss. 
We train our model for 120 epochs on a server with NVIDIA GeForce RTX 2080 Ti GPUs.  

\begin{table*} \small
	\centering
	\setlength\extrarowheight{2pt}
	\caption{Performance evaluation on iPER and ComplexMotion datasets. We quantitatively evaluate the performance on two scenarios: Image Reconstruction and Motion Imitation. $\uparrow$ indicates  higher is better and $\downarrow$ indicates  lower is better.}
	\label{compare}
	\resizebox{0.99\textwidth}{!}{
		\begin{tabular}{c|ccc|ccc|ccc|ccc} 
			\hline
			\multirow{3}{*}{\textbf{Methods}} & \multicolumn{6}{c|}{\textbf{ComplexMotion}}                                                                                                                                                                   & \multicolumn{6}{c}{\textbf{iPER}}                                                                                                                                                                              \\ 
			\cline{2-13}
			& \multicolumn{3}{c|}{Image Reconstruction}                                                    & \multicolumn{3}{c|}{Motion Imitation}                                                                         & \multicolumn{3}{c|}{Image Reconstruction}                                                    & \multicolumn{3}{c}{Motion Imitation}                                                                           \\ 
			\cline{2-13}
			& \multicolumn{1}{c|}{SSIM$\uparrow$} & \multicolumn{1}{c|}{PSNR$\uparrow$} & LPIPS$\downarrow$ & \multicolumn{1}{c|}{FID$\downarrow$} & \multicolumn{1}{c|}{IS$\uparrow$} & \multicolumn{1}{l|}{TCM$\uparrow$} & \multicolumn{1}{c|}{SSIM$\uparrow$} & \multicolumn{1}{c|}{PSNR$\uparrow$} & LPIPS$\downarrow$ & \multicolumn{1}{c|}{FID$\downarrow$} & \multicolumn{1}{c|}{IS$\uparrow$} & \multicolumn{1}{l}{TCM $\uparrow$}  \\ 
			\hline
			EDN \cite{chan2019everybody}                               & 0.823                               & 24.36                               & 0.061             & 64.12                                & 3.411                             & 0.534                              & 0.852                               & 24.48                               & 0.086             & 57.52                                & 3.305                             & 0.591                               \\
			FSV2V \cite{wang2019few}                             & 0.748                               & 22.51                               & 0.132             & 99.11                                & 3.164                             & 0.575                              & 0.824                               & 21.18                               & 0.108             & 107.29                               & 3.136                             & 0.754                               \\
			PoseWarp \cite{balakrishnan2018synthesizing}                          & 0.711                               & 21.42                               & 0.149             & 78.21                                & 3.109                             & 0.334                              & 0.792                               & 22.16                               & 0.119             & 115.23                               & 3.095                             & 0.601                               \\
			LWGAN \cite{liu2019liquid}                             & 0.789                               & 24.27                               & 0.081             & 85.30                                & 3.398                             & 0.683                              & 0.843                               & 22.32                               & 0.091             & 76.38                                & 3.258                             & 0.729                               \\
			C2F-FWN \cite{wei2021c2f}                           & 0.878                               & 25.68                               & 0.048             & 53.19                                & 3.408                             & 0.689                              & 0.847                               & 24.32                               & 0.074             & 60.12                                & 3.412                             & 0.769                               \\ 
			FakeMotion \cite{liu2022copy}                       & 0.883                               & 27.15                               & 0.040             & 48.03                                & 3.543                             & 0.773                              & 0.856                               & 25.86                               & 0.068             & 56.27                                & 3.461                             & 0.799                               \\
			\textbf{FakeVideo (Ours)}                              & \textbf{0.896}                      & \textbf{27.52}                      & \textbf{0.032}    & \textbf{46.62}                       & \textbf{3.728}                    & \textbf{0.813}                     & \textbf{0.868}                      & \textbf{26.72}                      & \textbf{0.049}    & \textbf{54.94}                       & \textbf{3.582}                    & \textbf{0.872}                      \\
			\hline
	\end{tabular}}
\end{table*}

\begin{table*}\small
	\centering
	\setlength\extrarowheight{2pt}
	\caption{Ablation study on different components of our method. Experiments are performed on iPER and ComplexMotion datasets. ``r/m X'' refers to removing X module in our network. The complete network consistently achieves the best results, which are highlighted.}
	\label{tab2}
	\resizebox{0.99\textwidth}{!}{
		\begin{tabular}{c|ccc|cc|ccc|cc} 
			\hline
			\multirow{3}{*}{\textbf{Methods}} & \multicolumn{5}{c|}{\textbf{ComplexMotion}}                                                                                                           & \multicolumn{5}{c}{\textbf{iPER}}                                                                                                                      \\ 
			\cline{2-11}
			& \multicolumn{3}{c|}{Image Reconstruction}                                                     & \multicolumn{2}{c|}{Motion Imitation}                 & \multicolumn{3}{c|}{Image Reconstruction}                                                     & \multicolumn{2}{c}{Motion Imitation}                   \\ 
			\cline{2-11}
			& \multicolumn{1}{c|}{SSIM$\uparrow$} & \multicolumn{1}{c|}{PSNR$\uparrow$} & LPIPS$\downarrow$ & \multicolumn{1}{c|}{FID$\downarrow$} & IS$\uparrow$   & \multicolumn{1}{c|}{SSIM$\uparrow$} & \multicolumn{1}{c|}{PSNR$\uparrow$} & LPIPS$\downarrow$ & \multicolumn{1}{c|}{FID$\downarrow$} & IS$\uparrow$    \\ 
			\hline
			\textbf{Our method, Complete}     & \textbf{0.896}                      & \textbf{27.52}                      & \textbf{0.032}    & \textbf{46.62}                       & \textbf{3.728} & \textbf{0.868}                      & \textbf{26.72}                      & \textbf{0.049}    & \textbf{54.94}                       & \textbf{3.582}  \\ 
			\hline
			\multicolumn{11}{c}{Ablation study on~\textbf{\textbf{Dense Skip Connections}}}                                                                                                                                                                                                                                                                  \\ 
			\hline
			r/m dense skip connections         & 0.868                               & 25.28                               & 0.050             & 62.39                                & 3.218          & 0.813                               & 24.21                               & 0.075             & 62.12                                & 3.271           \\ 
			\hline
			\multicolumn{11}{c}{Ablation study on~\textbf{\textbf{Self-Supervised Face Enhancement}}}                                                                                                                                                                                                                                                            \\ 
			\hline
			Image feature                     & 0.703                               & 22.42                               & 0.129             & 83.44                                & 3.215          & 0.634                               & 22.14                               & 0.108             & 99.34                                & 3.019           \\ 
			\hline
			2 face vectors                    & 0.728                               & 24.68                               & 0.129             & 78.20                                & 3.108          & 0.719                               & 21.34                               & 0.114             & 89.51                                & 3.167           \\
			3 face vectors                    & 0.758                               & 25.62                               & 0.079             & 56.80                                & 3.331          & 807                                 & 22.56                               & 0.089             & 73.33                                & 3.267           \\
			4 face vectors                    & 0.784                               & 26.08                               & 0.088             & 59.71                                & 3.304          & 0.753                               & 21.52                               & 0.098             & 75.28                                & 3.261           \\
			5 face vectors                    & 0.883                               & 27.15                               & 0.040             & 48.03                                & 3.543          & 0.856                               & 25.86                               & 0.068             & 56.27                                & 3.461           \\ 
			\hline
			1 candidate face                  & 0.732                               & 24.88                               & 0.139             & 75.20                                & 3.158          & 0.689                               & 20.24                               & 0.104             & 88.51                                & 3.067           \\
			2 candidate faces                 & 0.793                               & 26.22                               & 0.083             & 60.91                                & 3.371          & 0.746                               & 22.72                               & 0.088             & 75.54                                & 3.321           \\ 
			\hline
			\multicolumn{11}{c}{Ablation study on \textbf{Multiple Local GANs}}                                                                                                                                                                                                                                                                             \\ 
			\hline
			r/m multi-local GAN               & 0.872                               & 26.19                               & 0.053             & 61.49                                & 3.320          & 0.848                               & 24.19                               & 0.078             & 63.22                                & 3.373           \\
			\hline
	\end{tabular}}
\end{table*}

\subsection{Comparison with State-of-the-art Methods on Multiple Metrics (RQ1)}
In this section, we compare our proposed approach with existing state-of-the-art approaches, which include:
\begin{itemize}
	\item EDN (Everybody dance now) \cite{chan2019everybody}: A well-known pose-guided method for human motion copy, which makes amateurs dance like ballerinas.
	\item C2F-FWN (Coarse-to-fine flow warping network) \cite{wei2021c2f}: A novel motion copy method, which warps the layout based on the transformation flow.
	\item FakeMotion \cite{liu2022copy}: A motion copy approach, which generates human appearance with optimal transport theory and polishes the local body parts with multiple local GANs.
	\item FSV2V (Few-shot video2video) \cite{wang2019few}: A high-resolution and few-shot video generation method which is applicable to motion copy, facial expression transformation,  etc.
	\item LWGAN (Liquid warping GAN) \cite{liu2019liquid}: A unified warping framework which implements human motion copy, appearance (clothes) transfer, and novel view generation.
	\item PoseWarp \cite{balakrishnan2018synthesizing}: A motion copy method for sport scenes. In the method, 3D poses rather than 2D poses are utilized as the  motion intermediary, which provide the spatial characteristics of a motion.
\end{itemize}

To quantitatively compare the performance of our method and existing approaches, we divide the applications into two scenes: \emph{Image Reconstruction} and \emph{Motion Imitation}. For \emph{Image Reconstruction}, we perform self-mimicry experiments in which persons imitate actions from themselves. In other words, we feed the pose skeleton of a subject into the network and output the human image of the same subject. We adopt Structural Similarity (SSIM) \cite{wang2004image} as a low-level metric, Peak Signal to Noise Ratio (PSNR) and Learned Perceptual Image Patch Similarity (LPIPS) \cite{zhang2018unreasonable} as the perceptual level metrics to evaluate the quality of the generated image sequence.
For \emph{Motion Imitation}, we perform cross-mimicry where persons imitate the movements of others. Put differently, we input the pose skeleton of a subject into the network and output the human image of another subject.
We utilize the Inception Score (IS) \cite{salimans2016improved} and Frechet Inception Distance score (FID) \cite{heusel2017gans} to examine the differences between the generated images and the ground truth images. In addition, following the method in \cite{wei2021c2f}, we employ Temporal Consistency Metric \cite{yao2017occlusion} to measure the temporal continuity of the generated video. The experimental results on the \textit{ComplexMotion} and \textit{iPER} datasets are summarized in Table~\ref{compare}. From the table, we have the following observations. (1) Among the existing methods, C2F-FWN \cite{wei2021c2f} and FakeMotion \cite{liu2022copy} achieve the current state-of-the-art performance on both the two datasets. (2) \textit{FakeVideo} is able to outperform state-of-the-art approaches in both \textit{Image Reconstruction} and \textit{Motion Imitation}. For example, in image reconstruction and motion imitation, \textit{FakeVideo} gains 7.2\% and 12.4\% improvements on PSNR and FID metrics respectively. The significant performance improvements suggest the  potential of \textit{FakeVideo} to perform motion copy. The experimental results on the \textit{SoloDance} datasets are summarized in Table~\ref{solodance}. From the table, we have the following observations. (1) Among the existing methods, C2F-FWN [10] achieves the current state-of-the-art performance on SoloDance datasets. (2) \textit{FakeVideo} is able to outperform state-of-the-art approaches. For example, \textit{FakeVideo} gains 4.4\% and 4\% improvements on PSNR and FID metrics respectively. The performance improvements suggest the  potential of \textit{FakeVideo} to perform motion copy.

\begin{table}[h]
		
		\centering
		\caption{Performance evaluation on SoloDance datasets. }
		\label{solodance}
		\resizebox{0.5\textwidth}{!}{
		\begin{tblr}{
				cells = {c},
				vline{2-6} = {-}{},
				hline{1-2,9} = {-}{},
			}
			Methods                   & SSIM           & PSNR           & LPIPS          & FID            & TCM            \\
			EDN [6]                           & 0.811          & 23.22          & 0.051          & 53.17          & 0.347          \\
			FSV2V [7]                          & 0.721          & 20.84          & 0.132          & 112.99         & 0.106          \\
			PoseWarp [13]                  & 0.692          & 19.80          & 0.147          & 120.13         & 0.102          \\
			LWGAN [9]                      & 0.786          & 20.87          & 0.106          & 86.53          & 0.176          \\
			C2F-FWN [10]                    & 0.879          & 26.65          & 0.049          & 46.49          & 0.641          \\
			\textbf{FakeVideo (Ours)} & \textbf{0.893} & \textbf{27.82} & \textbf{0.038} & \textbf{44.72} & \textbf{0.739} 
		\end{tblr}}
\end{table}

\subsection{Visual Comparison with State-of-the-art Methods (RQ2)}
Further, we visualize the generated results of state-of-the-art approaches on three datasets, which are depicted in Fig.~\ref{fig7}. Empirically, PoseWarp \cite{balakrishnan2018synthesizing} and FSV2V \cite{wang2019few} may result in distorted body shapes and absent limbs. We conjecture the reasons are that they do not consider fusing information across multiple scales, leading to inevitable information dropping in the generation process. EDN \cite{chan2019everybody} achieves realistic visual results. However, the generated human faces of EDN usually have blurred facial parts. LWGAN \cite{liu2019liquid} and C2F-FWN \cite{wei2021c2f} could effectively copy the motions according to the optical flow, however, they have difficulties in generating fine-grained clothes and hairs. In contrast, our method yields a more realistic human body and plausible local details. 
More visual results of our method are demonstrated in Fig.~\ref{fig6}.  We see that our method consistently generates realistic frames.  

As shown in Fig.~\ref{fig7}, the first column (Target Person) illustrates the target person, the second column (Source Person) demonstrates the source person, the third column presents desired poses, which are obtained from videos of the source person (not the target person), and the remaining columns represent the generated frames of the target person. As shown in Fig.~\ref{fig7} and Fig.~\ref{fig6}, we would like to clarify that the source person and the target person are not the same individual, they are different in faces, body shapes, clothes, and even in gender. Fig.~\ref{fig7} and Fig.~\ref{fig6} contain multiple source and target persons pairs from three datasets.


\begin{table}\scriptsize
	\centering
	\setlength\extrarowheight{1.5pt}
	\caption{Ablation study on different loss functions. $\uparrow$ indicates higher is better, while $\downarrow$ indicates lower is better.}
	
	\begin{tabular}{c|c|c|c|c|c|c} 
		\hline
		& \multicolumn{3}{c|}{Complex Motion}               & \multicolumn{3}{c}{iPER}                           \\ 
		\cline{2-7}
		& SSIM$\uparrow$ & PSNR$\uparrow$ & LPIPS$\downarrow$ & SSIM$\uparrow$ & PSNR$\uparrow$ & LPIPS$\downarrow$  \\ 
		\hline
		$L_p$      & 0.838          & 25.10          & 0.058           & 0.820          & 24.11          & 0.081            \\
		$L_{GW}$     & 0.883          & 27.15          & 0.040           & 0.856          & 25.86          & 0.068            \\
		$L_p+L_{GW}$ & \textbf{0.896} & \textbf{27.52} & \textbf{0.032}  & \textbf{0.868} & \textbf{26.72} & \textbf{0.049}   \\
		\hline
	\end{tabular}
	\label{abcloss}
\end{table}

\begin{table}\scriptsize
	\centering
	\setlength\extrarowheight{1.5pt}
	\caption{Ablation study on the memory module over iPER and ComplexMotion datasets, where `w/o' represents removing (without) the memory module from our method. $\uparrow$ indicates higher is better, $\downarrow$ indicates lower is better.}
	\label{tab4}
	\begin{tabular}{c|c|c|c|c|c|c} 
		\hline
		& \multicolumn{3}{c|}{Complex Motion}               & \multicolumn{3}{c}{iPER}                           \\ 
		\cline{2-7}
		& SSIM$\uparrow$ & PSNR$\uparrow$ & LPIPS$\downarrow$ & SSIM$\uparrow$ & PSNR$\uparrow$ & LPIPS$\downarrow$  \\ 
		\hline
		w/o memory & 0.892          & 27.38          & 0.036           & 0.860          & 26.38          & 0.051            \\
		with memory & \textbf{0.896} & \textbf{27.52} & \textbf{0.032}  & \textbf{0.868} & \textbf{26.72} & \textbf{0.049}   \\
		\hline
	\end{tabular}
\end{table}

\subsection{Study on Key Components of \textit{FakeVideo} (RQ3)}
In this subsection, we study the effects of the different components of our method. With this goal in mind, we tried (1) removing dense skip connections from the pose-to-appearance generation GAN, (2) utilizing different kinds of loss functions in the generation network, (3) removing memory module from our framework, (4) using different face enhancement strategies, and (5) removing multiple local GANs which are responsible for local enhancement. 

\textbf{Removing dense skip connections from the pose-to-appearance GAN.} We first investigate the effect of removing the \textit{dense skip connections} in the pose-to-appearance GAN. From Table~\ref{tab2}, we observe that the performance of the method degrades significantly upon the removal of dense skip connections. This is consistent with our intuition that the dense skip connections could integrate multi-level latent features and is able to access lower-level pose details, contributing to better performance.

\textbf{Utilizing different kinds of loss functions.} Then, we examine the effects of different kinds of loss functions. In the pose-to-appearance generation GAN, we employ a perceptual loss $\mathcal{L}_p$ and a Gromov-Wasserstein Loss $\mathcal{L}_\textit{GW}$. To study the impacts of the two loss functions, we conduct comparing experiments on using $\mathcal{L}_p$, $\mathcal{L}_\textit{GW}$, and the combined losses $\mathcal{L}_p + \mathcal{L}_\textit{GW}$, respectively. The empirical results are elaborated in Table~\ref{abcloss}. The combined losses (\textit{i.e.,} $\mathcal{L}_p + \mathcal{L}_\textit{GW}$) achieves the best performance, while using $\mathcal{L}_\textit{GW}$ alone yields better results than using $\mathcal{L}_p$ alone.

\textbf{Removing memory module from our framework.}  In order to evaluate the effectiveness of our memory module, we further try removing the episodic memory module from our framework. As shown in Table~\ref{tab4}, with the memory module, our approach achieves 0.14 and 0.24 higher PSNR scores than its counterpart without the memory module on the two datasets. These evidences show that the memory module does play an important role in boosting the generation quality of our method.

\textbf{Using different face enhancement strategies.} For self-supervised face enhancement, we consider two schemes to select similar face images for the generated face: facial similarity computed using VGG image features and using the proposed face vector field. The results are demonstrated in Table~\ref{tab2}. Empirically, the face vector field strategy achieves significantly better performance, which is in accordance with our intuition. The image features of faces actually emphasize the appearance information (\emph{i.e.}, colors and eye shapes) while ignoring face orientation information, 
which has difficulties in effectively selecting similar face images to enhance the generated face. Our face vector field strategy is capable of accurately representing more fine-grained face orientation details, which is conducive to selecting similar face images that are more valuable to compensate for the facial details of the generated face. 

We also examine the influence on the number of face vectors, as shown in Table~\ref{tab2}. We observe that the quality of the generated image gradually increases with the growing number of face vectors. 
We also ablate the number of similar face images selected, as shown in Table~\ref{tab2}. We find that three images might provide sufficient facial information, resulting in informed self-supervised face enhancement. 

\begin{figure}[t]
	\centering
	\includegraphics[width=\columnwidth]{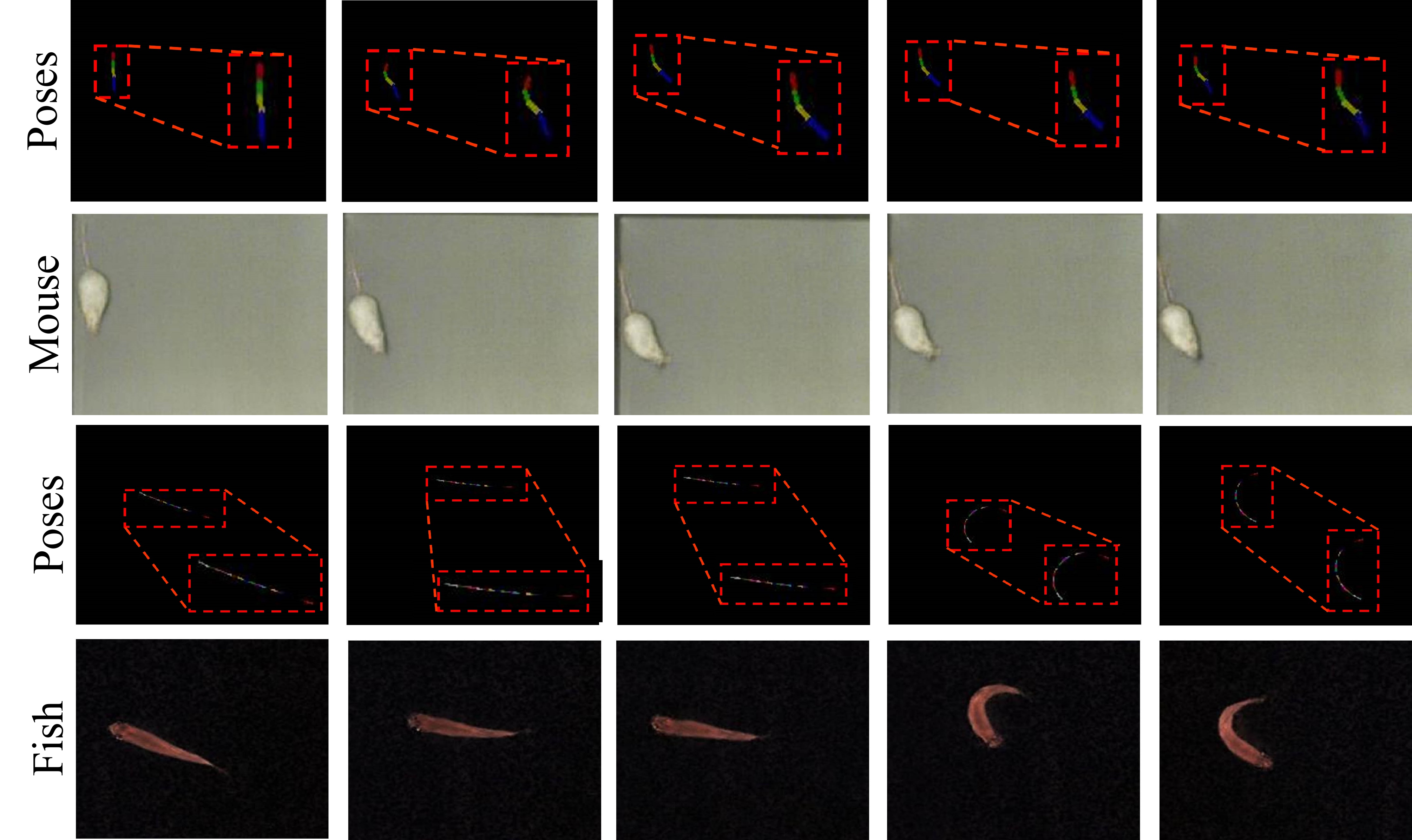}
	\caption{Results on generating mice and fish videos. In the figure, we have enlarged the poses of mice and fish. Please zoom in to see the details.}
	\label{fig8}
\end{figure}

\textbf{Removing local GANs from the local enhancement module.} Finally, we remove the multiple local GANs to examine their contributions. As shown in Table~\ref{tab2},  FID significantly increases from $48.03$ to $61.49$ upon the removal of the local GANs. This dramatic image quality degeneration highlights the effectiveness of the local GANs in the local refinement. Particularly, the residual images of human body parts generated by the  local GANs reveal the difference in color and texture details between the generated image of the body part and the ground truth, facilitating the generation of more lifelike local images.
\subsection{Motion Copy on Other Articulated Objects (RQ4)}
In addition to performing motion copy on humans, we are curious whether our approach could be generalized to other articulated objects including zebra fish and mouse. To this end, we further conduct experiments on \textit{fish} \cite{xu2017lie} and \textit{mouse} \cite{liu2022investigating} datasets. Interestingly, our method could be adapted to copy motions of fish and mouse. Empirical results are demonstrated in Fig.~\ref{fig8}. Take fish as an example, in the training stage, we first employ Lie-X \cite{xu2017lie} to detect the desired poses of the fish from given videos. Then, we disentangle the frames of the video of the target fish into foreground and background using Mask-RCNN \cite{he2017mask}. Thereafter, we feed the desired poses into our pose-to-appearance generation network, where the network architecture remains the same but with the feature size adapted to fit fish. We would like to point out that we do not enhance the details of the generated frames, since a zebra fish is small in body size and the pose-to-appearance generation seems to be sufficient to yield realistic fish frames. Finally, we couple the generated foreground and the background, obtaining the entire fish videos. In the inference stage, the network is fed with the desired poses from another fish and we could synthesize a lively video of the target fish, where the target fish swims and acts like another fish. Experiments on fish and mouse show that our method is able to copy motions of other articulated objects.

\subsection{Discussion about the Computational Time Comparison}\quad
	Motion transfer models could be classified into two categories: dedicated-purpose  models and general-purpose models. Specifically, the dedicated-purpose models excel in generating fake videos of a specific person and offer high video quality at the expense of longer training time. In contrast, general-purpose models have the capability to generate fake videos of any person, necessitating less training time but yielding less satisfactory generation results compared to dedicated-purpose models. In this paper, we concentrate on dedicated-purpose models. We would like to emphasize that despite the longer training time required for our dedicated-purpose model, it offers a shorter inference time. Empirical results are represented in Table \ref{Time}.	 Specifically, during the inference phase, EDN \cite{chan2019everybody} achieves an average Frames Per Second (FPS) of 14.29, \cite{liu2022copy} achieves an average FPS of 15, and our method has an average FPS of 25.25.

\begin{table}[h]
	
	\centering
	\caption{Computational time Comparison.}
	\label{Time}
	\begin{tabular}{|c|c|c|c|} 
		\hline
		Methods & EDN   & FakeMotion & \textbf{FakeVideo (Ours)}  \\ 
		\hline
		FPS     & 14.29 & 15         & \textbf{25.25}           \\
		\hline 
	\end{tabular}
\end{table}

\subsection{User study}
\begin{figure}[ht]
	\centering
	\includegraphics[width=\columnwidth]{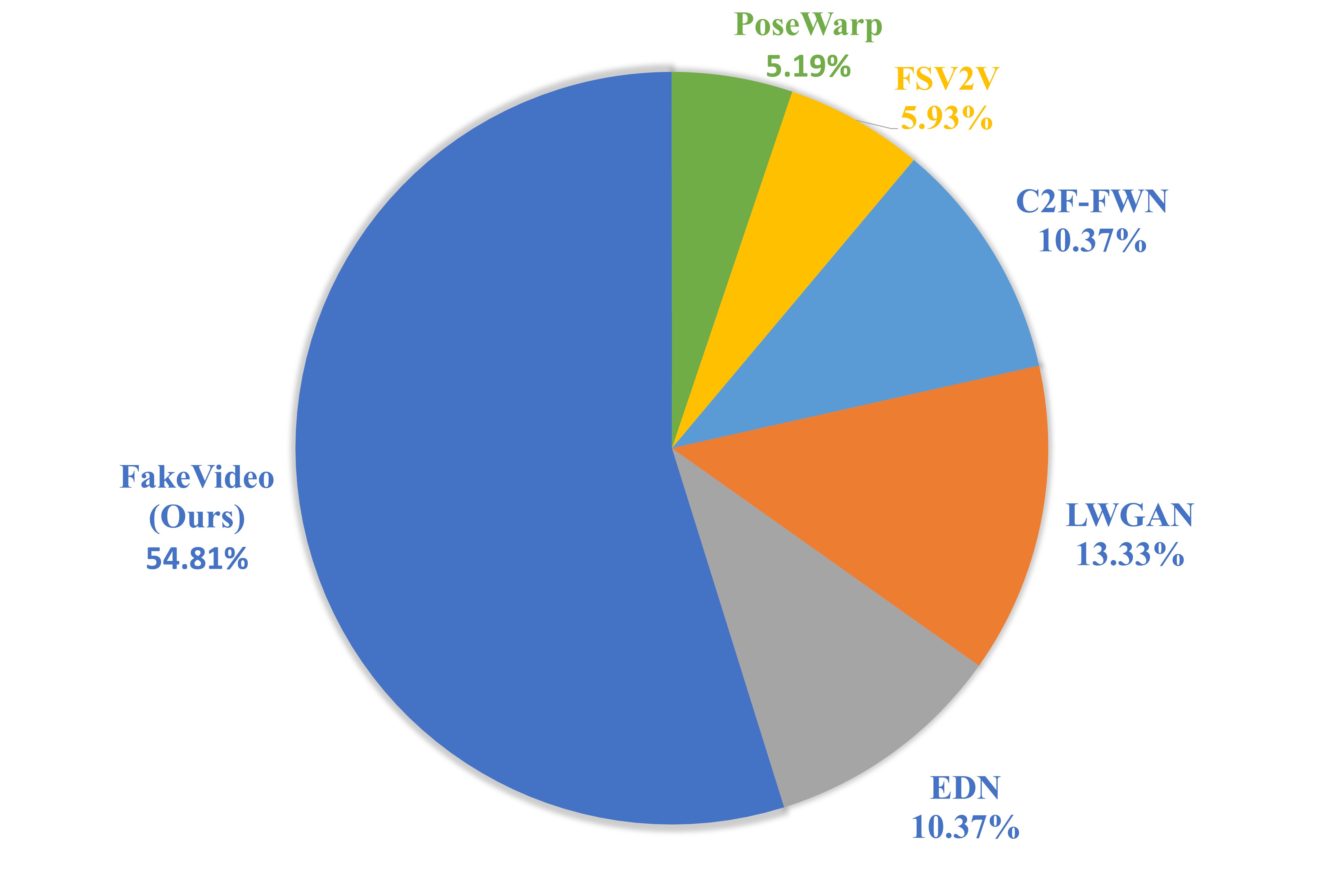}
	\caption{Result of the user study.}
	\label{user}
\end{figure}

We conduct a user study, engaging a cohort of 25 volunteers, to meticulously evaluate the quality of the generated results. Each participant is presented with six clusters of generated results, and subsequently requested to discern and designate the best results within each group. Ultimately, we collect a total of 25 responses, the results are shown in Fig. \ref{user}. We can see that the proposed FakeVideo gets the highest rating and significantly outperforms other methods (EDN \cite{chan2019everybody}, PoseWarp \cite{balakrishnan2018synthesizing}, FSV2V \cite{wang2019few}, C2F-FWN \cite{wei2021c2f}, and LWGAN \cite{liu2021liquid}).

\subsection{Failure Case Analysis}

\begin{figure}[t]
	\centering
	\includegraphics[width=\columnwidth]{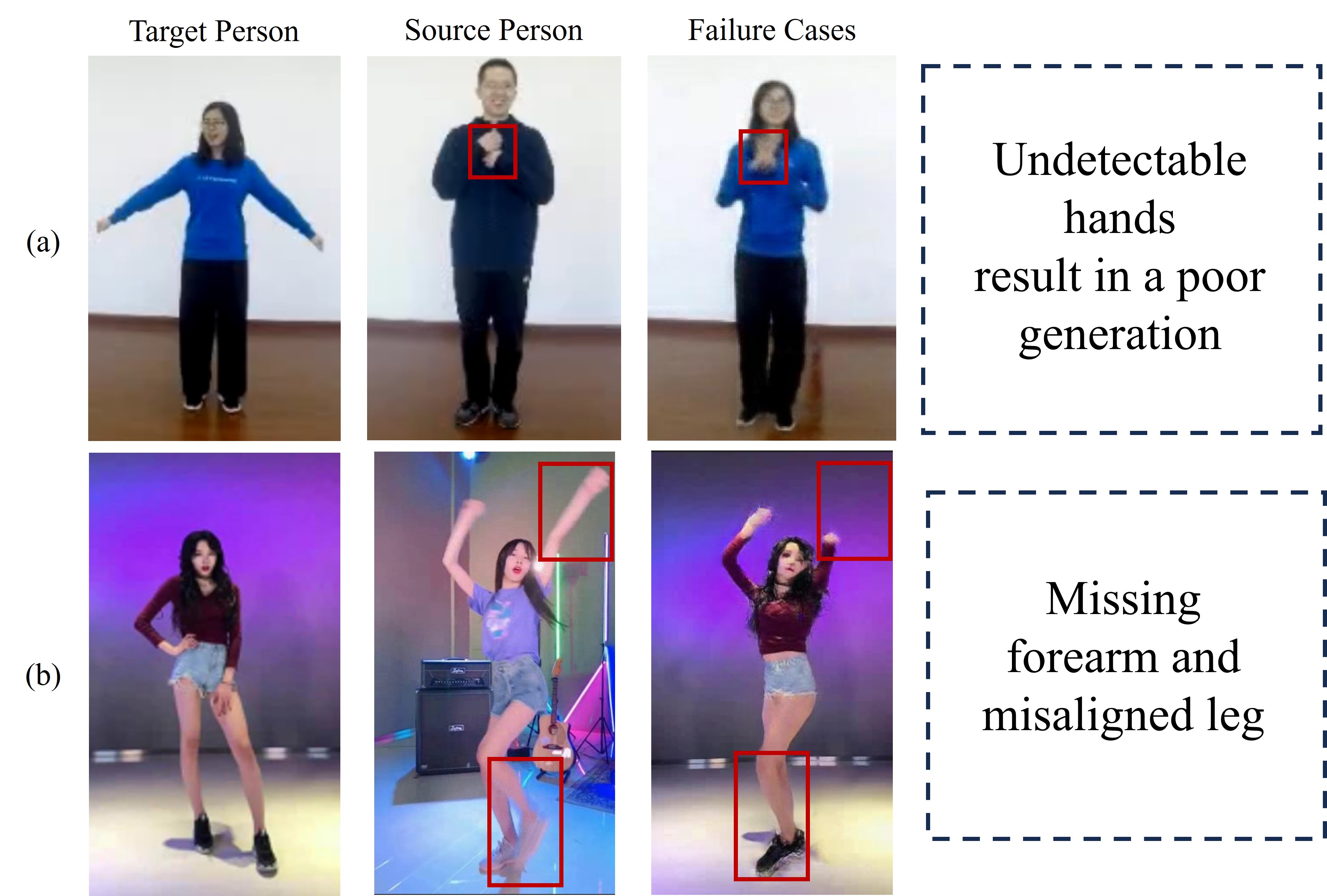}
	\caption{The failure cases of our method.}
	\label{case}
\end{figure}

	Interestingly, we also observed a few failure cases. The failure cases are shown in Figure \ref{case}. The first failure case is shown in Fig. \ref{case} (a). When the hands of the source character are undetectable, the generated hands are not realistic enough. The second failure case is shown in Fig. \ref{case} (b). If there are artifacts in the input source image, such as elongated arms and lower legs, this will result in a target image with missing forearms and misaligned lower legs.
	In the subsequent work, we plan to implement the following measures to further improve the generation:

(1) We will develop a more accurate human pose estimation framework, which plays an important role in the task of human motion copy.

(2) Additionally, we will enhance the network structure. Specifically, in cases where a generated limb of the synthesized frame is missing, the network will strive to generate a limb that aligns with the target person, thereby ensuring a more coherent output.




\section{Conclusion}
\label{conclusion}

In this work, we present a novel approach \textit{FakeVideo} for motion copy. The crucial ingredient is proposing a pose-to-appearance generation network with  Gromov-Wasserstein and perceptual losses, and a memory module that consistently learns from its past poor generations. We further introduce a self-supervised face enhancement module that resorts to face frames with similar orientations to polish facial details of the generated face. 
Interestingly, our approach could be generalized to other articulated objects, including fish and mouse. Extensive empirical results on five datasets,  \emph{iPER, ComplexMotion, SoloDance, fish and mouse datasets}, demonstrate the efficacy of the proposed method.

\footnotesize
\bibliographystyle{IEEEtran}
\bibliography{TDSC}
%

\begin{IEEEbiography}[{\includegraphics[width=1in,height=1.25in,clip,keepaspectratio]{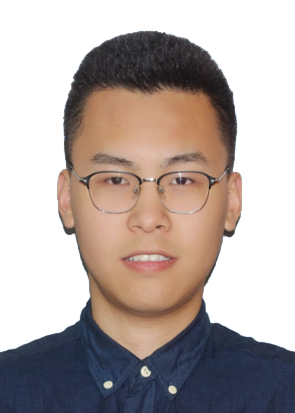}}]{Sifan Wu} is currently pursuing a ph.D. degree at Jilin University. He received his B.E. and M.E. degrees from Hebei GEO University  and Zhejiang Gongshang University, respectively. His research interests include computer vision, motion copy, and pose estimation.
\end{IEEEbiography}

\begin{IEEEbiography}[{\includegraphics[width=1in,height=1.25in,clip,keepaspectratio]{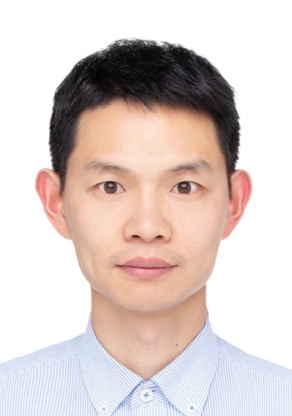}}]{Zhenguang Liu}
is currently a research professor of  Zhejiang University. He had been a research fellow  in National University of Singapore and A*STAR.  (Agency for Science, Technology and Research, Singapore).  He respectively received his Ph.D. and B.E.  degrees from Zhejiang University and Shandong  University, China.	  His research interests include multimedia data analysis and smart contract security. Various parts  of his work have been published in first-tier venues  including PAMI, ACM CCS, CVPR, ICCV, TKDE, TIP, WWW, TDSC, AAAI, ACM MM, INFOCOM, IJCAI, etc. Dr. Liu has served as technical program  committee member for top-tier conferences such as CVPR, ICCV, WWW,  AAAI, IJCAI, ACM MM, session chair of ICGIP, local chair of KSEM, and  reviewer for IEEE PAMI, IEEE TVCG, IEEE TPDS, IEEE TIP, ACM TOMM, IEEE MM,  etc. 
\end{IEEEbiography}

\begin{IEEEbiography}[{\includegraphics[width=1in,height=1.25in,clip,keepaspectratio]{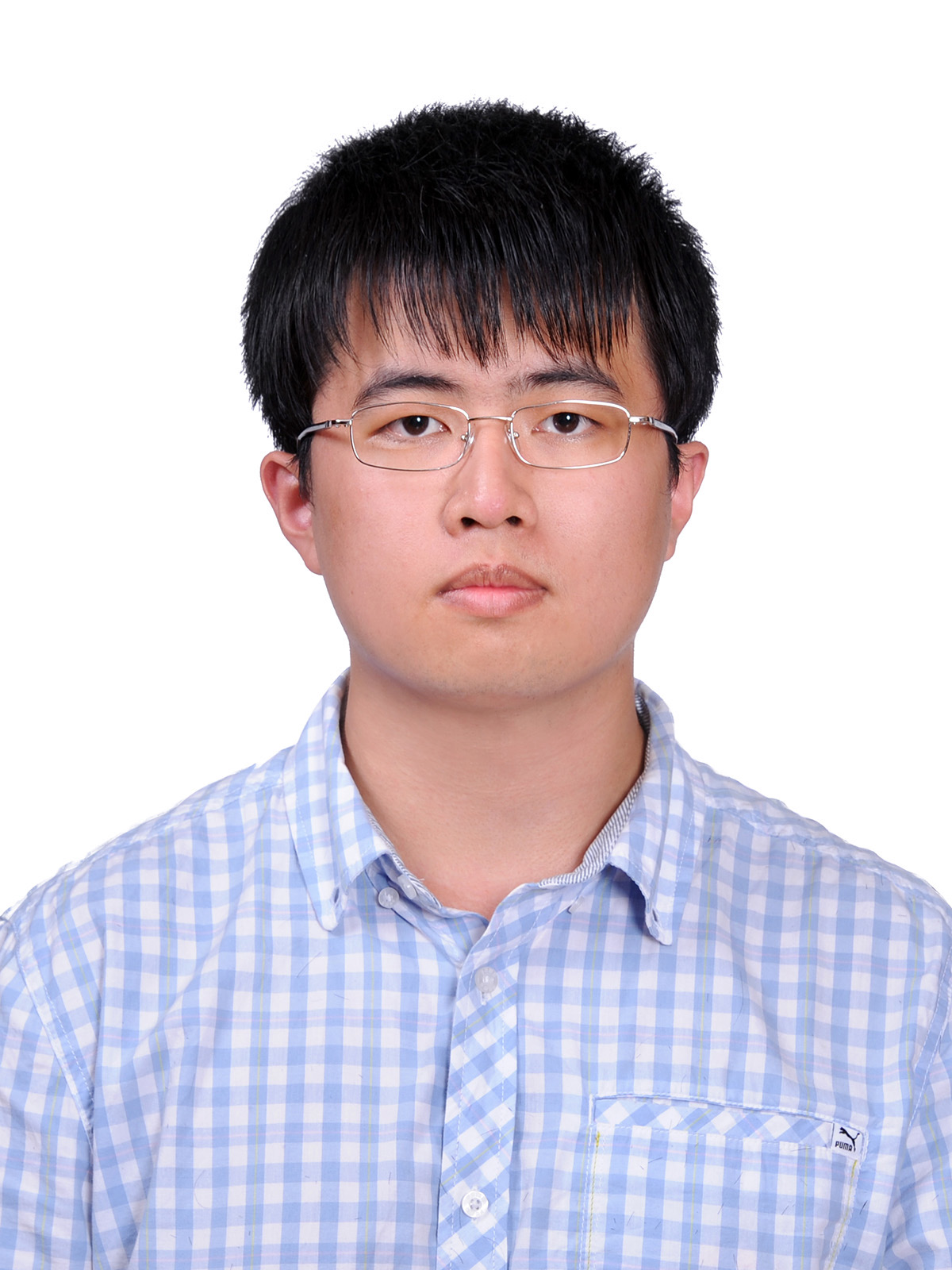}}]{Beibei Zhang} received his B.Sc. (Hons) degree in Information Technology from The Hong Kong Polytechnic University in 2017 and his M.Eng. degree from the Department of Electrical \& Computer Engineering, University of Toronto, in 2020. He is currently a Research Engineer in Zhejiang Lab, Hangzhou, China. His research interests include distributed systems, cloud computing, and peer-to-peer networks.
\end{IEEEbiography}

\begin{IEEEbiography}[{\includegraphics[width=1in,height=1.25in,clip,keepaspectratio]{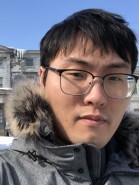}}]{Zhongjie Ba} received the Ph.D. in Computer Science and Engineering from the State University of New York at Buffalo, USA, in 2019. He is currently a Professor with the School of Cyber Science and Technology, College of Computer Science and Technology, Zhejiang University, China. He was a Postdoctoral Researcher in the School of Computer Science at McGill University, Canada. His current research interests include the security and privacy aspects of Internet of Things, forensic analysis of multimedia content, and privacy-enhancing technologies in the context of collaborative deep learning. Results have been published in peer reviewed top conferences and journals, including CCS, NDSS, INFOCOM, ICDCS, and IEEE Trans. Inf. Forensics Security. Currently, Zhongjie Ba serves as an Associate Editor of IEEE Internet of Things Journal and the technical program committee of several conferences in the field of Internet of Things and wireless communication.
\end{IEEEbiography}

\begin{IEEEbiography}[{\includegraphics[width=1in,height=1.25in,clip,keepaspectratio]{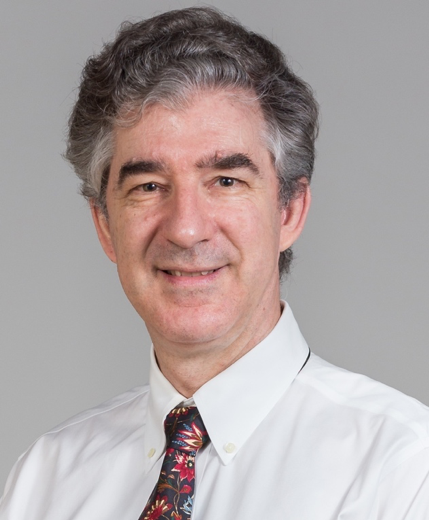}}]{Roger Zimmermann}  (M’93–SM’07) received the M.S. and Ph.D. degrees from the University of
	Southern California, Los Angeles, USA, in 1994	and 1998, respectively. He is currently an Associate Professor with the Department of Computer Science,	National University of Singapore (NUS), Singapore, where he is also the Deputy Director with the Smart Systems Institute, and co-directed the Centre of Social Media Innovations for Communities. He has co-authored a book, seven patents, and over 200 conference publications, journal articles, and	book chapters. His research interests include streaming media architectures, distributed systems, mobile and geo-referenced video management, collaborative environments, spatio-temporal information management, and mobile location-based services. He is a Senior Member of the IEEE and a Distinguished Member of the ACM.
\end{IEEEbiography}
\begin{IEEEbiography}[{\includegraphics[width=1in,height=1.25in,clip,keepaspectratio]{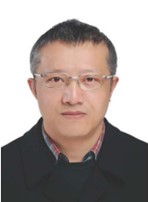}}]{Xiaosong Zhang}  received the B.S. degree in dynamics engineering from Shanghai Jiao Tong University, Shanghai, China, in 1990, and the M.S. and Ph.D. degrees in computer science from the University of Electronic Science and Technology of China (UESTC), Chengdu, China, in 2011.,He is currently a Professor with the School of Computer Science and Engineering, UESTC. He has worked on numerous projects in both research and development roles. These projects include device security, intrusion detection, malware analysis, software testing, and software verification. He has coauthored a number of research articles on computer security. His current research interests include software reliability, software vulnerability discovering, software test case generation, and reverse engineering.
\end{IEEEbiography}
\begin{IEEEbiography}[{\includegraphics[width=1in,height=1.25in,clip,keepaspectratio]{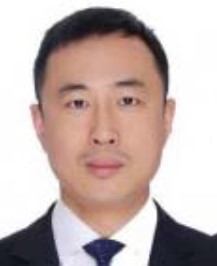}}]{Kui Ren} (Fellow, IEEE, Fellow, ACM) received the PhD degree from Worcester Polytechnic Institute. He is currently a professor and an associate dean with the College of Computer Science and Technology, Zhejiang University, where he also directs the Institute of Cyber Science and Technology. Before that, he was the SUNY Empire Innovation professor with the State University of New York at Buffalo. He has authored or coauthored extensively in peer-reviewed journals and conferences. His research interests include data security, IoT security, AI security, and privacy. His h-index is 74 and the total publication citation exceeds 32000 according to Google Scholar. He was the recipient of Guohua Distinguished Scholar Award from ZJU in 2020, IEEE CISTC Technical Recognition Award in 2017, SUNY Chancellor's Research Excellence Award in 2017, Sigma Xi Research Excellence Award in 2012, NSF CAREER Award in 2011,Test-of-time Paper Award from IEEE INFOCOM, and many best paper awards from the IEEE and ACM, including the MobiSys'20, ICDCS'20, Globecom'19, ASIACCS'18, and ICDCS'17. He is a distinguished member of the ACM and a clarivate highly-cited researcher. He is a frequent reviewer of funding agencies internationally and was on the editorial boards of many IEEE and ACM journals. He is the chair of the SIGSAC of ACM China.
\end{IEEEbiography}

\end{document}